\documentclass[ 
                    author={Jinhyun Jason Park},
                supervisor={Dr. Rui Ponte Costa},
                    degree={BSc},
                     title={Representations learnt by SGD and Adaptive learning rules},
                  subtitle={Conditions that Vary Sparsity and Selectivity in Neural Networks},
                      type={Research},
                      year={2021} ]{dissertation}
                      
\newcommand{\comb}[2]{{}_{#1}\mathrm{C}_{#2}}

\begin{document}





\maketitle


\frontmatter


\makedecl


\tableofcontents
\listoffigures
\listoftables
\listofalgorithms



\chapter*{Acknowledgements}

\noindent
Thank you very much to Nayeon Jenna Song, Rui Ponte Costa, and Heng Wei. Special thanks to Will Greedy.


%

\mainmatter
\chapter{Abstract}
From the point of view of the human brain, continual learning can perform various tasks without mutual interference
\cite{flesch2018comparing}. An effective way to reduce mutual interference can be found in sparsity and selectivity of neurons.  
According to Aljundi et al. and Hadsell et al. \cite{aljundi2018selfless, hadsell2020embracing}, imposing sparsity at the representational level\footnote{Sparsity in terms of neuron activations. Hoyer's sparsity metric \cite{hoyer2004non} can measure this.} is advantageous for continual learning because sparse neuronal activations encourage less overlap between parameters, resulting in less interference.
Similarly, highly selective neural networks are likely to induce less interference since particular response in neurons will reduce the chance of overlap with other parameters. Considering that the human brain performs continual learning over the lifespan, finding conditions where sparsity and selectivity naturally arises may provide insight for understanding how the brain functions.
This paper investigates various conditions that \textit{naturally} increase sparsity and selectivity in a neural network. 
This paper tested different optimizers with Hoyer's sparsity metric \cite{hoyer2004non} and CCMAS selectivity metric \cite{morcos2018importance} in MNIST classification task. 
It is essential to note that investigations on the \textit{natural} occurrence of sparsity and selectivity concerning various conditions have not been acknowledged in any sector of neuroscience nor machine learning until this day. 
This paper found that particular conditions increase sparsity and selectivity such as applying a large learning rate and lowering a batch size. 
In addition to the relationship between the condition, sparsity, and selectivity, the following will be discussed based on empirical analysis: 1. The relationship between sparsity and selectivity and 2. The relationship between test accuracy, sparsity, and selectivity.

\chapter{Introduction}
This chapter introduces the structural summary of this paper. The core contents of this paper are divided into four main branches: (1) Contextual Background, (2) Technical Background, (3) Results, and (4) Conclusion. 
\newline 
\newline 
\textbf{Contextual Background} consists of 3 sections: (1) Prior work, (2) The relationship between sparsity, selectivity, and continual learning, and (3) A brief explanation of why 4 optimizers: SGD, Adagrad, Adadelta and Adam were chosen. (2) explains the definition of sparsity and selectivity and how these are related to continual learning in detail.
\newline 
\newline 
\textbf{Technical Background} provides a detailed description of each optimizer because it provides valuable insights such as how parameter updating rules could be related to sparsity and selectivity. 
It has been found that the sparsity and selectivity tend to increase or decrease depending on how much past gradients are taken into account in SGD.  
Secondly, it explains the correlation between the batch size and generalization ability. 
Thirdly, Hoyer's sparsity metric \cite{hoyer2004non} was used for calculating sparsity and CCMAS metric \cite{morcos2018importance} was used for calculating selectivity; thus, the algorithm of these metric will be explained in detail. Furthermore, this chapter will clarify why the two metrics were chosen instead of other metrics. 
\newline 
\newline 
The \textbf{Results} chapter is mainly subdivided into each experiment of condition where a section describes one condition, and its performance. Each section explains experimental details, result, analysis, and conclusion. The result of the experiment will be compared with the baseline model, which is introduced at the beginning of the \textbf{Results} chapter. Also, considering the fact that sparsity and selectivity can be measured on both training-set and testing-set, the difference between them will be examined. The following conditions have been experimented: varying the number of hidden layers, varying the hyper-parameters, varying the batch size, varying the number of neurons in the hidden layer, varying the diversity of classes in the batch, applying a convolutional layer, and comparing sorted vs unsorted data-set. 
Test accuracy, sparsity, and selectivity are examined for each condition. 
Not only considering a single condition into account when measuring performance, but two conditions will also be combined in section \ref{combinationofconditionslabel}. 
This experiment is conducted to answer the following question: if two conditions with high performance are incorporated, will the combined model show better performance than the others? 
Lastly, based on the extensive data regarding test accuracy, sparsity, and selectivity investigated by changing conditions, this paper empirically analyses how these attributes are related to each other.
\newline 
\newline 
The \textbf{Conclusion} chapter summarises all findings from the previous chapters and sections. Instead of simply summarising the findings, this paper states factors that affect sparsity and selectivity in bullet point format. It also introduces possible further works and limitations of this project.

\chapter{Contextual Background} \label{chap:context}
\section{Prior Researches}
Many previous studies examined the relationship between neural network, sparsity, and selectivity. 
L1-regularisation (LASSO regularisation) was applied to weight matrices to reduce the connection between the neurons in order to increase sparsity \cite{hebiri2020layer, lee2007efficient}. 
Gale et al. \cite{gale2019state} used the following three techniques: magnitude pruning \cite{zhu2017prune}, variational dropout \cite{molchanov2017variational}, and L0-regularization \cite{louizos2017learning} to induce sparsity in deep neural network\footnote{They used Transformer trained on WMT 2014 English-to-German and ResNet-50 trained on ImageNet for learning tasks.}. 
Dey et al. \cite{dey2019pre} presented the pre-defined sparse neural network where weights between hidden layers are not fully connected; instead, partly connected. The connection between neurons is \textit{structured} where every node in a layer has the same number of connections going to the right and coming from the left. 
In terms of selectivity, Leavitt et al. \cite{leavitt2020selectivity} researched the relationship between CCMAS selectivity \cite{morcos2018importance} and test accuracy. They directly applied a regularisation technique to the CCMAS selectivity metric and analysed its effect.
These researches used particular techniques that directly affect sparsity and selectivity of the neural network. On the other hand, this paper focuses on the \textit{natural occurrence} of sparsity and selectivity by using various conditions. 
\section{Sparsity, Selectivity, and Continual Learning}
Sparsity is one of the most trending topics in machine learning and neuroscience. It provides better performance in terms of efficiency while maintaining the performance of the neural network, such as saving computational resources and preventing over-fitting \cite{hebiri2020layer}. According to Sharkey et al. and Murre, \cite{sharkey1995analysis, murre2014learning}, sparsity appears to play an important role in encouraging selective feature response; therefore, sparsity and selectivity seem closely related. 
\newline 
\newline 
\textbf{Sparsity} means that relatively few neurons in a population respond to a particular stimulus \cite{willmore2001characterizing}. 
For example, assume there are 100 neurons that are connected to a single neuron A. When A fires, only 10 out of 100 neurons responds. The response of neurons can be related to the activation value of neurons in the neural network. 
The sigmoid activation function is used, making the activation value from 0 to 1. 
If the activation value of a specific neuron is closer to 0, it is less active. Conversely, if it is close to 1, it is more active.
It is possible to calculate the sparsity of neural network by using activation values and Hoyer's metric (A detailed algorithm of calculating Hoyer's metric will be discussed in section \ref{sparsityandselectivitylabel}). If the value obtained through the metric is large, hidden layer neurons of the neural network are highly sparse (The majority of neurons are not active). On the contrary, if the value obtained through the metric is small, hidden layer neurons of the neural network are barely sparse (The majority of neurons are active).
\newline 
\newline 
\textbf{Selectivity} means that a neuron reacts only to certain features and do not respond unless they are. Features can be colour, class, shape, etc. For example, class feature selectivity can be described by assuming: there are 100 neurons and there are two classes X and Y. 
The first 50 neurons respond to X, and the last 50 neurons only respond to Y.
Then, the activation of neurons is perfectly selective. 
In practice, neurons that respond to a particular class are not strictly fixed. 
Their activations overlap.
This paper uses class selectivity (also can be called digit selectivity as each digit corresponds to each class) metric. 
Similar to sparsity, selectivity can be measured using the activation values of neurons and CCMAS selectivity metric. 
If the value obtained through the metric is large, the network are highly selective. On the contrary, if the value obtained through the metric is small, the network are hardly selective.
\newline 
\newline 
These two properties play a vital role in the brain; because high sparsity and
selectivity are likely to imply that neurons are processing the information efficiently. Also, the higher the sparsity and selectivity, the more similar to continual learning of the human brain. 
Many studies revealed that the activation of neurons is sparse and selective in various areas of the brain.
It is widely known that the dentate gyrus implements pattern separation\footnote{The ability of the brain to separate the pattern is directly related with classification task.} by sparsifying the input from the entorhinal cortex \cite{chavlis2017dendrites, treves1994computational}. 
Sparsifying the input reduces overlap between neurons when a stimulus is presented to the brain. This is likely to encourage less interference and forgetting enabling continual learning; since there will be less chance where similar task-related parameters can be overlapped \cite{hadsell2020embracing}. 
Likewise, Barth and Poulet \cite{barth2012experimental} proposed that neurons fires sparsely in the neocortex and Barlow \cite{barlow1972single} stressed that the brain should act sparsely for efficient information representation.
Furthermore, it was found that the activation of neurons is sparse through various measurement methods \cite{olshausen2004sparse, hulme2014sparse, foldiak2003sparse}.
Besides sparsity, it has been proven through many papers that the activation of neurons is selective to certain features. 
Selective neurons reduce overlap between neurons resulting in less overlap between parameters. This is likely to encourage less interference, encouraging continual learning.
Hubel and Wisel \cite{hubel1962receptive} revealed that a cat's primary visual cortex is orientation selective. 
Likewise, Levick \cite{levick1967receptive} found that in the rabbit retina, there are orientation-selective ganglion cells. 
Bowers et al. \cite{bowers2014neural} argued that selective representations support the co-activation of features such as words, objects, and faces which is vastly related to continual learning in short term memory. 
It can be seen that neurons in the brain are sparse and selective for sensory and perceptual stimulus. 
As this paper discusses under what conditions sparsity and selectivity arise naturally, it is likely to suggest how neurons are structured or how neurons process information in the brain. 
As a result, this research would provide insight into continual learning. Moreover, it is known that the human brain is sparse and selective in the visual cortex (V1). This study would help to understand how V1 functions as well.

\section{The Reasons behind choosing 4 optimizers}
\label{section21}
This project used SGD optimizer (with momentum) \cite{sutskever2013importance} and three adaptive learning rules: Adagrad \cite{duchi2011adaptive}, Adadelta \cite{zeiler2012adadelta}, and Adam \cite{kingma2014adam} optimizers. 
\newline 
\newline 
SGD is one of the most basic optimizers. Recent research has shown that SGD performs better in generalization ability than other adaptive optimizers \cite{wilson2017marginal, zhou2020towards}. Also, momentum hyper-parameter, which could be related to sparsity and selectivity, can be easily applied to SGD optimizer in Pytorch. 
I expected that the momentum would affect them since it is directly related to a parameter update rule.
\newline 
\newline 
Adagrad is one of the state-of-art adaptive learning rule optimizers. It is an optimizer that does not use a constant learning rate. The learning rate changes for each step and the different learning rate is applied for each parameter. For features (in images) that occur frequently, a low learning rate is applied. Conversely, for features (in images) that do not happen frequently, a high learning rate is applied. For these reasons, it is believed that Adagrad performs well when when dealing with sparse data \cite{ruder2016overview}. These characteristics of Adagrad was thought to be highly related to sparsity and selectivity; therefore, it has been decided to consider Adagrad for adaptive learning methods. The detailed definition of Adagrad will be discussed in chapter \ref{chap:technical}.
\newline 
\newline 
Adadelta is another popular adaptive gradient optimizer. The biggest downside of Adagrad is that the parameter update does not work very well as learning progresses. Adadelta solves this shortcoming of Adagrad. This phenomenon cannot be clearly explained without analysing the mathematics. This will be further discussed in chapter \ref{chap:technical}. Adadelta is an optimizer that does not need a learning rate. That is, the optimizer itself automatically adjusts the learning rate for each step. This feature seems to be related to the human brain because human beings do not set a learning rate when learning or perceiving objects. It also has a $\rho$ (rho) value that determines how much past information (gradient) affects the current step. These characteristics of Adadelta would be related to sparsity and selectivity of neurons; thus, it has been decided to consider Adadelta for adaptive learning methods.
\newline 
\newline 
Adam is also one of the most well-known adaptive gradient optimizers. Adam optimizer is an optimizer that combines SGD with momentum and RMSprop (RMSprop is an optimizer that overcomes the shortcoming of Adagrad) \cite{ruder2016overview}. In other words, Adam has two hyper-parameters $\beta_1$ and $\beta_2$ which are similar to momentum $\gamma$ in SGD with momentum and rho $\rho$ in RMSprop. These will be further discussed in chapter \ref{chap:technical}. These hyper-parameters are expected to affect sparsity and selectivity. According to Kingma and Ba \cite{kingma2014adam}, Adam works well than other adaptive optimizers in practice. Goodfellow et al. \cite{backprop} mentioned that although Adam's learning rate should be tuned from the default value, it is generally robust to the choice of hyper-parameters. 
It is meaningful to analyse sparsity and selectivity for one of the best and renowned optimizer. For these reasons, it has been decided to consider Adam.

\chapter{Technical Background} \label{chap:technical}
This chapter covers two main topics. Firstly, it will explain algorithms to obtain sparsity and selectivity metric. It is crucial to identify which sparsity and selectivity metric were used in this paper because there are various ways to calculate them. It also explains why these metrics were chosen. Secondly, four optimizers which were mentioned in section \ref{section21} will be described in detail. A detailed description of each optimizer would help to analyse the result in chapter \ref{chapter4}. It may be possible to use the algorithm of the optimizers to relate the occurrence of varying sparsity and selectivity. In addition, the relationship between the parameter update rule and batch size is described in detail for section \ref{varyingbatchsize}.
\newpage
\section{The Calculation of Sparsity and Selectivity}
\label{sparsityandselectivitylabel}
This paper evaluates sparsity using Hoyer’s metric and selectivity by the CCMAS metric. 
According to Hurley and Rickard \cite{hurley2009comparing}, various models calculate sparsity; however, Hoyer’s metric was chosen because its utility outperforms the other sparsity metrics \cite{hurley2009comparing}. 
The CCMAS metric was chosen because the metric is inspired by neuroscience, widely-used, and easy to calculate \cite{leavitt2020selectivity}. 
The calculation of sparsity and selectivity in this section assumes that there is only one hidden layer and \textit{batch size} = 50.
\subsection{Sparsity}
\begin{algorithm}[H]
\SetAlgoLined
    sparsities = list( )\\
    all\_epochs\_sparsity = list( )\\
    \For{epoch \textbf{in} range(30)}{
    activations = list( )\\
        \For{training}{
            do training\\
        }
        \For{testing}{
            do testing\\
            // model.layer\_activations contains \\
            // activation values of each neuron in the hidden layer.\\
            activations.append(model.layer\_activations)    // \textbf{Step1}\\
        }
        \underline{sparsities}.append(activations)\\
    }
    \For{sparsity \textbf{in} sparsities}{
        \underline{stacked\_sparsity} = torch.stack(sparsity)\\
        reshape stacked\_sparsity to 10000 $\times$ 256    // \textbf{Step2}\\
        N $\leftarrow$ a number of neurons in the hidden layer\\
        c $\leftarrow$ stacked\_sparsity\\
        S $\leftarrow$ calculated values of Hoyer's sparsity by using N and c // \textbf{Step3}\\ 
        Avg\_S $\leftarrow$ torch.mean(S) // \textbf{Step4}\\
        \underline{all\_epochs\_sparsity.append(Avg\_S)} // \textbf{Step5}\\
    }
\caption{Sparsity calculation}
\end{algorithm}
\vspace{0.5cm}
\noindent \textbf{Step1}: Accumulate all activation values\footnote{Activation values could be acquired by using \textit{register\_forward\_hook} in Pytorch. The hook is registered at the sigmoid activation function; therefore, the activation value bound between 0 and 1.} from neurons in the hidden layer.
The shape of \underline{sparsities} is (total iterations) $\times$ (batch size) $\times$ (number of neurons in the hidden layer) =
200 $\times$ 50 $\times$ 256
\vspace{1mm}
\newline 
\textbf{Step2}: Reshape \underline{stacked\_sparsity} to 10000 $\times$ 256.
\vspace{1mm}
\newline 
\textbf{Step3}: Calculate Hoyer’s sparsity metric by setting $N$ = (number of neurons in the hidden layer) = 256 and $c$ = \underline{stacked\_sparsity}. $j$ denotes $j$-th neuron in the hidden layer. The shape of ${\sum_{j}^{} c_j}$ is 10000 $\times$ 1; therefore, there are 10000 samples of calculated sparsity. 
\begin{equation} 
    \textit{Hoyer's sparsity} = (\sqrt{N} - \frac{ {\sum_{j}^{} c_j}} {\sqrt{ {\sum_{j}^{} {c_j}^2}}})(\sqrt{N} - 1)^{-1}
\end{equation}
\newline 
\textbf{Step4}: Return the average of all values acquired in \textbf{Step3}. This step averages a sparsity of 10000 testing images.
\vspace{1mm}
\newline 
\textbf{Step5}: \underline{all\_epochs\_sparsity} contains all sparsity values for each epoch from 1 to 30.
\newpage

\subsection{Selectivity}
\begin{algorithm}[H]
\SetAlgoLined
    all\_epochs\_sparsity\_avg = list( )\\
    all\_epochs\_sparsity\_std = list( )\\
    \For{epoch \textbf{in} range(30)}{
        all\_vals = list( )\\
        activations = [ \{ 0:[ ], 1:[ ], 2:[ ], 3:[ ], 4:[ ], 5:[ ], 6:[ ], 7:[ ], 8:[ ], 9:[ ] \} for x in range(256) ]\\
        \For{training}{
            do training\\
        }
        \For{testing}{
            do testing\\
            \For{activation, label \textbf{in} model.layer\_activations, labels}{
                \For{i \textbf{in} range(256)}{
                    \underline{activations}[ i ][ label ].append(activation[ i ]) // \textbf{Step1}\\
                }
            }
        }
        \underline{final\_activation} = [ dict( ) for x in range(256) ]\\
        \For{i, neuron \textbf{in} enumerate(\underline{activations})}{
            \For{k, v \textbf{in} neuron.items( )}{
                \underline{final\_activation[ i ][ k ]} $\leftarrow$ sum(v) / len(v) // \textbf{Step2}
            }
        }
        \For{vals \textbf{in} \underline{final\_activation}}{
            $u_{max}$ $\leftarrow$ maximum of vals \\
            $u_{-max}$ $\leftarrow$ summation of vals - $u_{max}$ \\
            selectivity $\leftarrow$ calculated value of CCMAS selectivity by using $u_{max}$ and $u_{-max}$ // \textbf{Step3} \\
            all\_vals.append(selectivity)
        }
        // \textbf{Step4}\\
        avg\_all\_vals $\leftarrow$ average of all\_vals \\ 
        std\_all\_vals $\leftarrow$ standard deviation of all\_vals \\
        // \textbf{Step5}\\
        \underline{all\_epochs\_sparsity\_avg}.append(avg\_all\_vals)\\
        \underline{all\_epochs\_sparsity\_std}.append(std\_all\_vals)\\
    }
\caption{Selectivity calculation}
\end{algorithm}
\vspace{0.5cm}
\noindent \textbf{Step1}: Accumulate all activation values from neurons in the hidden layer. 
The shape of the \underline{activations} is (number of neurons in the hidden layer) x (number of classes) = 256 x
10.
For example, \underline{activations[ 0 ]} denotes $0$-th neuron in the hidden layer. It contains a list of activation values for each digit
(dictionary form): \underline{activations[ 0 ]} = $\{ 0:L_0, 1:L_1, ... , 9:L_9  \}$ where $L_i$ denotes a list of activation values for digit
$i$.
\vspace{1mm}
\newline 
\textbf{Step2}: Average all activation values for each digit. For instance, \underline{final\_activation[ 0 ]} = $\{0:Avg_0, 1:Avg_1, ..., 9:Avg_9\}$
where $Avg_i$ denotes the average of $L_i$.
\vspace{1mm}
\newline 
\textbf{Step3}: Calculate the selectivity value of each neuron by using the CCMAS metric. A few experiments encountered nan values; thus, $\epsilon = 10^{-7}$ is added to the denominator to solve the issue.
\newline 
\begin{equation} 
    \textit{CCMAS selectivity} = \frac{ u_{max} - u_{-max} }{ u_{max} + u_{-max} + \epsilon}
\end{equation} 
where
\begin{align*}
    u_{max} &= maximum(Avg_i)\\
    u_{-max} &= \frac{{\sum_{i=0}^{9} Avg_i - u_{max}}}{9}
\end{align*}
\textbf{Step4}: Calculate the average and the standard deviation of all neurons' selectivity.
\vspace{1mm}
\newline 
\textbf{Step5}: \underline{all\_epochs\_sparsity\_avg} and \underline{all\_epochs\_sparsity\_std} contain all selectivity values for each epoch.

\newpage

\section{4 Optimizers}  \label{optimizerdeepexplain}
\subsection{SGD and Introduction to Gradient Descent Algorithm}
\label{batchsizerelationship}
\begin{figure}[h!]
  \centering
  \includegraphics[width=0.8\textwidth]{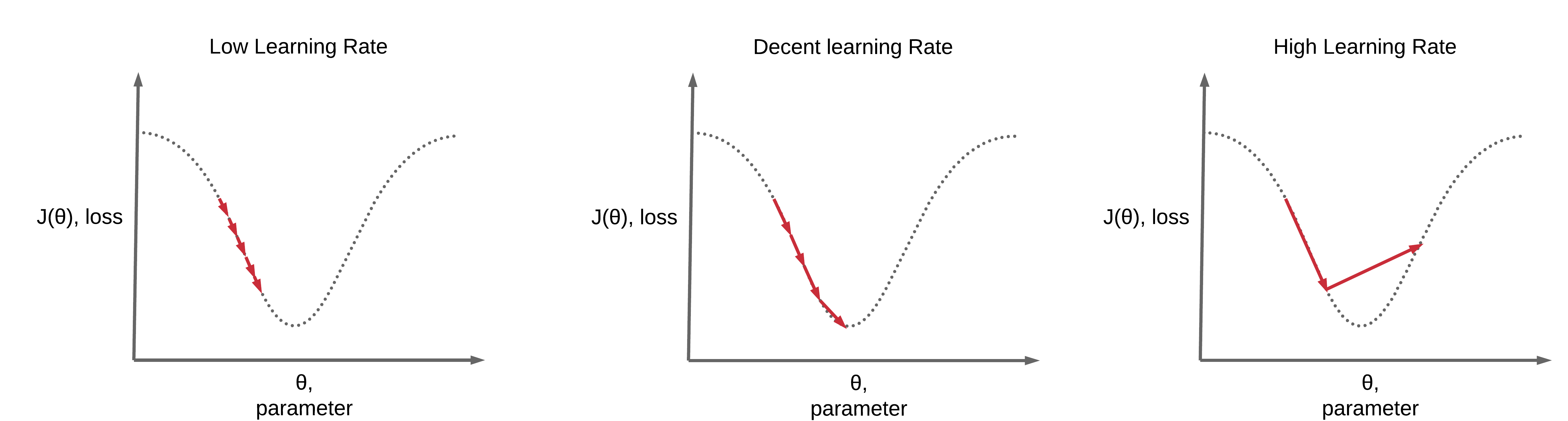}
  \caption{Learning rate \cite{learningrate}.}
  \label{learning_rate_image}
\end{figure}

Gradient descent is a method to find the minimum value of the objective function $J(\theta)$ parameterised by $\theta$. After finding the slope of $J(\theta)$, update $\theta$ in the opposite direction of the slop. This can be expressed as the following equation:
\begin{equation} \label{eq1}
\theta_{t+1} = \theta_t - \eta \cdot \nabla_\theta J(\theta_t)
\end{equation}
Equation \ref{eq1} is the basic form of SGD algorithm. $\eta$ (eta) is a learning rate hyper-parameter that determines how much to update the parameter (also called as weight) $\theta$. 
The learning rate should be set before the training. 
If the learning rate is set too low, it will take a long time to train the model because a step of the parameter update is small. 
Assuming there is a time limitation or epoch limitation, it may not be possible to reach the local optimum of the objective function $J(\theta)$. 
On the other hand, if the learning rate is set too high, which means the parameter update is too large, then it may fail to find the local optimum since learning would be very unstable. 
However, it is known that sometimes having a high learning rate would provide a chance to escape from sharp minimum \cite{setting2018jordan} (The explanation about sharp and flat minimum will be described in this chapter).
Figure \ref{learning_rate_image} explains the effect of a small and large learning rate. It is common to aim for a `Descent learning rate', which eventually arrives at the local optimum better than the other two.
\newline 
Usually, it is common to have multiple parameters. In equation \ref{eq2}, $i$ is taken into account, which means the equation can deal with numerous parameters.
\begin{equation} \label{eq2}
    \theta_{t+1,i} = \theta_{t,i} - \eta \cdot \nabla_\theta J(\theta_{t,i})
\end{equation}
It is important to differentiate between gradient descent algorithm, stochastic gradient descent algorithm and mini-batch gradient descent algorithm because they are used interchangeably. 
Also, the clarification of these methods would provide better understanding when analysing the relationship between the batch size and parameter updating rule.
\newline 
\newline 
\textbf{Gradient descent} is used to update parameters for the entire training data set. This can be expressed by the following equation:
\begin{equation} 
\theta_{t+1,i} = \theta_{t,i} - \eta \cdot \nabla_\theta J(\theta_{t,i})
\end{equation} 
\textbf{Stochastic gradient descent} is used to update parameters for each sample in the training set. This is the same as setting \textit{batch size} = 1 when running SGD algorithm. When updating parameters, it is unstable to reach the local optimum rather than the gradient descent because only one example is used for parameters update. However, it has been proven that the stochastic gradient descent algorithm would also stably converge to the local optimum like gradient descent \cite{ruder2016overview}. In addition, due to its instability, it may move to a potentially better local minimum \cite{ruder2016overview}. The following equation can express stochastic gradient descent:
\begin{equation} 
\theta_{t+1,i} = \theta_{t,i} - \eta \cdot \nabla_\theta J(\theta_{t,i}; x^{a}, y^{a})
\end{equation}
\textbf{Mini-batch gradient descent} is an algorithm that utilises the advantage of gradient descent algorithm and stochastic gradient descent algorithm. Instead of updating parameters by looking at all data or a single example, parameters are updated by looking at $m$ examples. This can be expressed by the following equation:
\begin{equation} \label{minibatcheq}
\begin{split}
    \theta_{t+1,i} = \theta_{t,i} - \eta \cdot \nabla_\theta J(\theta_{t,i}; x^{a:a+m}, y^{a:a+m})
\end{split}
\end{equation}
\subsection{Batch size and generalization ability} \label{batch size and GA}
It seems like the mini-batch gradient descent is used merely to change the batch size.
However, there is a notable difference when the batch size is small and when the batch size is large.
According to \cite{masters2018revisiting}, the small batch size has two main advantages over the large batch size. Firstly, generalization ability can be improved by using smaller batch training. Secondly, training stability can be enhanced. In other words, training is possible with a wider range of learning rate. Re-writing the definition of \ref{minibatcheq} (These equations below are derived from Masters and Luschi \cite{masters2018revisiting}.),
\begin{equation} \label{eq3}
\begin{split}
    \theta_{k+1} & = \theta_{k} + \eta \cdot \Delta \theta_k\\
    \Delta \theta_k & = -\frac{1}{m} {\sum_{i=1}^{m} {\nabla_{\theta} L_i (\theta_k)} }
\end{split}
\end{equation}
where $t$ has been changed to $k$ and $J$ has been changed to $L$. Also, $a$ has been removed for clarity. Using the equations in \ref{eq3}, the mean value of weight update could be expressed by the following:
\begin{equation} \label{eq4}
E[\ \eta \Delta \theta ]\ = - \eta E[\ \nabla_\theta L(\theta) ]\
\end{equation}
This equation \ref{eq4} is for m examples (Remember \textit{batch size} = m in equation \ref{minibatcheq}). For a single example, $m$ should be divided by each side of the equation:
\begin{equation} \label{eq5}
\frac{1}{m} E[\ \eta \Delta \theta ]\ = - \frac{\eta}{m} E[\ \nabla_\theta L(\theta) ]\
\end{equation}
In order to maintain the left-hand side of the equation \ref{eq5} (mean of SGD parameter update per training sample) to be constant, $\frac{\eta}{m}$ should be constant despite increasing or decreasing the batch size. In other words, the learning rate $\eta$ should be changed depending on $m$. This is known as linear scaling, where the learning rate increases as the batch size increases. The linear scaling rule is a widely used technique \cite{krizhevsky2014one, chen2016revisiting, bottou2018optimization, smith2017don, jastrzkebski2017three}.
Considering these facts, equation \ref{eq3} would be re-written by the following equation:
\begin{equation} \label{eq6}
\begin{split}
    \theta_{k+1} = \theta_{k} - \alpha {\sum_{i=1}^{m} {\nabla_{\theta} L_i (\theta_k)} }
\end{split}
\end{equation}
where $\alpha = \frac{\eta}{m}$. By using this equation \ref{eq6}, batch size $m$, and $n$ steps of weight update, the parameter update would be expressed by the following:
\begin{equation}
    \theta_{k+n} = \theta_{k} -\alpha {\sum_{j=0}^{n-1} {\sum_{i=1}^{m} \nabla_{\theta} L_{i+jm} (\theta_{k+j})}}
\end{equation}
By increasing the batch size by $n$ times, the following parameter update equation would be expressed. Note that the left hand side of the equation is not $\theta_{k+n}$, it is $\theta_{k+1}$ ($k+\frac{n}{n} = k+1$).
\begin{equation}
    \theta_{k+1} = \theta_{k} -\alpha {\sum_{i=1}^{nm} \nabla_{\theta} L_{i} (\theta_{k})}
\end{equation}
By comparing these two equations, it is possible to see that if the batch size is small, the latest gradient can be used for updating parameters. 
This property leads the neural network to be more generalized \cite{masters2018revisiting}.
In contrast, if the batch size is large, the latest gradient cannot be used and the old gradient is used for the parameter update, leading to low generalization ability.
Many studies show that small batch is efficient for generalization and optimization convergence \cite{wilson2003general, lecun2012efficient, keskar2016large}. In a neural network, objective function (also known as a cost function) $L$ is very complicated and non-convex. Non-convex means there are many local optima. 
If the function is non-convex and highly complex, the gradient can be hugely affected by the parameter update; therefore, if the algorithm uses an old gradient, it is likely that the parameter update will not be accurate. 
There is an another paper that supports this argument. 
According to Keskar et al. \cite{keskar2016large}, the larger the batch size, the higher the probability of convergence to the sharp minimum of $L$, which degrades the generalization performance. 
\begin{figure}[h!]
  \centering
  \includegraphics[width=0.6\textwidth]{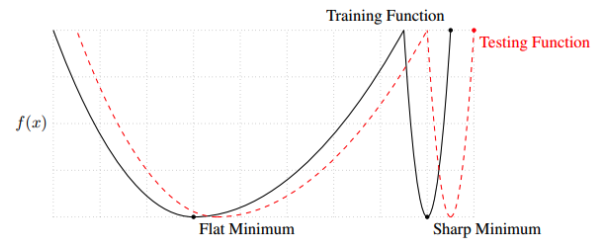}
  \caption{Flat minimum and Sharp minimum \cite{keskar2016large}}
  \label{flatvssharpminimum}
\end{figure}
Keskar et al. and Hochreiter and Schmidhuber \cite{keskar2016large, 1997generalisation} stated that, when using a small batch, the gradient noise increases in the step computation; thus, it can easily escape from the sharp minimum and converge to the flat minimum, which implies better generalization performance\footnote{There is a paper by Dinh et al. \cite{dinh2017sharp} that contradicts this theory. They argued that there is no relationship between the shape of the local optimum and generalization ability. As the aim of this research is not focused on this topic, I will not discuss the content in depth.}. 
Assuming an objective function $f(x)$ looks like Figure \ref{flatvssharpminimum}, it seems reasonable that sharp minimum would be sensitive to the parameter update than flat minimum. 
These two papers \cite{masters2018revisiting, keskar2016large} could not mathematically prove why the small batch converges to flat minimum and why flat minimum is suitable for generalization. However, they empirically proved the relationship between the batch size and the generalization ability. 
In conclusion, generalization ability would likely to be increased if the small the batch size is used rather than the large batch size. 
\subsection{SGD with momentum}
\label{sgdmomentum}
SGD with momentum \cite{sutskever2013importance} is an advanced method of SGD. As the name implies, momentum (a.k.a. inertia) is added to SGD. Adding momentum may allow the algorithm to escape from the local optima by using the force of inertia. In addition, it may allow the algorithm to escape from the saddle point. This is often explained as rolling the ball on the objective function $J$. 
If the ball starts at a high place, it would be possible to escape from the shallow local optimum or saddle point due to inertia. SGD with momentum can be expressed by the following equation.
\begin{equation}
\theta_{t+1,i} = \theta_{t,i} - v_{t,i}
\end{equation}
where
$$v_{t,i} = \gamma \cdot v_{t-1,i} + \eta \cdot \nabla_\theta J(\theta_{t,i})$$
$$v_{-1,i} = 0$$
The parameter update takes the velocity $v$ into account. The velocity is the sum of the learning rate multiplied by the current gradient and the $\gamma$ (momentum hyper-parameter) multiplied by the previous velocity. If $\gamma = 0$, then the algorithm becomes exactly the same as SGD algorithm. If $\gamma \neq 0$, then the algorithm takes the velocity into account. In general, $\gamma$ ranges from 0 to 1. The closer it is to 1, the more it takes the velocity into account. By analysing this equation, \textit{recent past} gradient is expected to have relatively greater influence on the parameter update than \textit{far past} gradient; because the velocity decays with rate $\gamma$.
Pytorch takes a slightly different form. The learning rate is multiplied outside of the velocity function. SGD with momentum in the Pytorch version could be expressed by the following:
\begin{equation}
    \theta_{t+1,i} = \theta_{t,i} - \eta \cdot v_{t,i}
\end{equation}
where
$$v_{t,i} = \gamma \cdot v_{t-1,i} + \nabla_\theta J(\theta_{t,i})$$
$$v_{-1,i} = 0$$
Another thing to note is that the larger the $\gamma$, the more the SGD with momentum fluctuates when training. According to Smith et al. \cite{smith2017don}, a scale of random fluctuation of SGD with momentum could be expressed by the following:
\begin{equation} \label{sgdmomentumeq}
    g = \frac{\epsilon}{1-\gamma} (\frac{N}{B} - 1)
\end{equation}
where $N$ denotes the size of the data-set and $B$ denotes the batch size. It can be also induced that a smaller batch size increases fluctuation. 
\subsection{Adagrad}
\label{adagradinthepaper}
The fatal drawback of SGD is that the same learning rate applies equally to all parameters. Also, the learning rate always keeps the same value until the training is finished. Adagrad complements this problem. The learning rate of Adagrad varies from parameter to parameter and step to step. Adagrad has the following assumption. 
Firstly, parameters have been updated by significant value means that the current position got closer to the local minimum. Since the position is close to the local minimum, step size should be decreased. In other words, a lower running rate is applied to the parameter update that has been updated frequently. 
Secondly, parameters have been updated by a small value means that the current position did not get closer to the local minimum yet. Step size should not be largely decreased\footnote{This means that the value of decrements should not be large. Adagrad step size always decreases for every iteration.}. In other words, a higher learning rate is applied to the parameter update that has been updated infrequently. Adagrad algorithm \cite{duchi2011adaptive} could be expressed by the following:
\begin{equation}
\theta_{t+1,i} = \theta_{t,i} - \frac{\eta}{\sqrt{G_{t,i}}+\epsilon} \odot \nabla_\theta J(\theta_{t,i})
\end{equation}
where:
$$G_{t,i} = G_{t-1,i} + \nabla_\theta J(\theta_{t,i}) \odot \nabla_\theta J(\theta_{t,i})$$
$$G_{-1,i}=0$$
$\odot$ refers to Hadamard product \cite{horn1990hadamard}. By looking at the equation, it is possible to derive that the step size decreases as learning continues.

\subsection{RMSProp}
\label{RMSPropinthepaper}
This optimizer is explained because it will be used for the explanation of Adam.
One of the major drawbacks of Adagrad is that $G$ increases infinitely. This means that if epoch or training lasts a long time, the parameter update will occur with a very small step. RMSProp complements this drawback of Adagrad. The equation of RMSProp can be expressed by the following equation:
\begin{equation}
\theta_{t+1,i} = \theta_{t,i} - \frac{\eta}{\sqrt{G_{t,i}}+\epsilon} \odot \nabla_\theta J(\theta_{t,i})
\end{equation}
where:
$$G_{t,i} = \gamma G_{t-1,i} + (1 - \gamma) \cdot \nabla_\theta J(\theta_{t,i}) \odot \nabla_\theta J(\theta_{t,i})$$
$$G_{-1,i}=0$$
By looking at the above equation, $\gamma$ and $(1 - \gamma)$ can be observed. This is also known as the moving average technique. $\gamma$ prevents $G$ from increasing to infinity. In addition, by using $\gamma$, the degree of taking past gradient and present gradient can be determined. The larger the $\gamma$, more attention is paid to previous gradient. The smaller the $\gamma$, more attention is paid to the current gradient. The past gradient decays at rate $\gamma$; thus, \textit{recent past} gradients will have a relatively greater influence on the parameter update than \textit{far past} gradients.

\subsection{Adadelta}
\label{Adadeltarho}
In RMSProp, $\gamma$ was introduced to complement the shortcoming of Adagrad. In Adadelta, the way to acquire $G$ is the same as in RMSProp. This can be expressed as follows.
\begin{equation}
E[g^2]_{t,i} = \rho E[g^2]_{t-1,i} + (1-\rho)g^2_{t,i}
\end{equation}
where $G_{t,i}$ is replaced by $E[g^2]_{t,i}$. And $\nabla_\theta J(\theta_{t,i})$ is replaced by  $g_{t,i}$. This new notation $E[g^2]_{t,i}$ is a method of expressing the exponential moving average interpreted as follows: $E[g^2]$ approximately average over the last $\frac{1}{1-\rho}$ $E[g^2]$ values. For instance, if $\rho$=0.9, then $E[g^2]_{t,i}$ approximately averages over the last $\frac{1}{1-0.9} = 10$ $E[g^2]$ values. If $\rho$=0.98, then $E[g^2]_{t,i}$ approximately average over last $\frac{1}{1-0.98} = 50$ $E[g^2]$ values. In other words, increasing $\rho$ is the same as increasing the average window size. This also means the optimizer adapts more slowly to the current gradient $g^2_{t,i}$ as $\rho$ increases. The parameter update formula is expressed by the following.
\begin{equation}
\Delta \theta_t = - \frac{\eta}{\sqrt{E[g^2]_{t,i} + \epsilon}} \odot g_{t,i}
\end{equation}
The author of Adadelta \cite{zeiler2012adadelta} mentioned that when updating a parameter, the left hand of the equation above has a unit of $\theta$, whereas the right hand side of the equation doesn't. To match the unit on both sides, the paper introduces a moving average of the parameter itself.
\begin{equation}
E[\Delta \theta^2]_{t-1,i} = \rho E[\Delta \theta^2]_{t-2,i} + (1-\rho)\Delta \theta^2_{t-1,i}
\end{equation}
The parameter update formula considering this is as follows. The equation satisfies $\theta$ unit on both the right and the left hand side.
\begin{equation}
\Delta \theta_{t,i} = - \frac{\sqrt{E[\Delta \theta^2]_{t-1,i} + \epsilon}}{\sqrt{E[g^2]_{t,i}+\epsilon}} \odot g_{t,i}
\end{equation}
By observing this equation, it can be seen that the learning rate is not needed. Zeiler \cite{zeiler2012adadelta} also stated that this algorithm does not require a manual tuning of the learning rate. However, Pytorch added a learning rate. The final formula of Adadelta can be expressed by the following:
\begin{equation}
\theta_{t+1,i} = \theta_{t,i} - \eta \cdot \frac{\sqrt{E[\Delta \theta^2]_{t-1,i} + \epsilon}}{\sqrt{E[g^2]_{t,i}+\epsilon}} \odot g_{t,i}
\end{equation}

\subsection{Adam}
\label{adambetas}
Adam is an algorithm that combines two methods: SGD with momentum and RMSProp. The `moving average' technique is applied to momentum. Adam algorithm would be expressed by the following.
\begin{equation}
\theta_{t+1} = \theta_t - \eta \cdot \frac{p_{t,i}}{\sqrt{q_{t,i}}+\epsilon}
\end{equation}
where:
$$p_{t,i} = \beta_1 p_{t-1,i} + (1-\beta_1)g_{t,i}$$
$$p_{-1,i} = 0$$
$$q_{t,i} = \beta_2 q_{t-1,i} + (1-\beta_2)g_{t,i}^2$$
$$q_{-1,i} = 0$$
$\beta_1$ is equal to $\gamma$ and $p_{t,i}$ is equal to $v_{t,i}$ in SGD with momentum. Likewise, $\beta_2$ is equal to $\rho$ and $q_{t,i}$ is equal to $G_{t,i}$ in RMSProp. 
By looking at the equation, it can be derived that if $\beta_1$ and $\beta_2$ are very large (usually $\beta_1$ and $\beta_2$ are close to 1) then $p_{t,i}$ and $q_{t,i}$ are too biased to 0 intially; thus, the parameter update would not work well. It can be adjusted by using the following equation.
\begin{align*}
    \hat{p}_{t,i} &= \frac{p_{t,i}}{1-\beta_1^t} \\
    \hat{q}_{t,i} &= \frac{q_{t,i}}{1-\beta_2^t}
\end{align*}
The adjustment can effectively solve the problem. The adjustment affects the parameter update a lot in the beginning. However, its effect gradually decreases as learning progresses. 
Taking this into consideration, the following equation is derived.
\begin{equation}
\theta_{t+1} = \theta_t - \eta \cdot \frac{\hat{p}_{t,i}}{\sqrt{\hat{q}_{t,i}}+\epsilon}
\end{equation}

\chapter{Results}
\label{chapter4}
\section{Experimental Details for baseline models}
\label{defaultsetting}
A hyper-parameter search was performed to find the best learning rate for four optimizers: SGD, Adagrad, Adadelta, and Adam.
SGD and Adam used $\eta \in \{0.1, 0.01, 0.001, 0.0001, 0.00001\}$. 
Adagrad and Adadelta used $\eta \in \{1.0, 0.1, 0.01, 0.001, 0.0001\}$. 
Other hyper-parameters were set as default with \textit{batch size} = 50. 
The split ratio of training and testing is 6 : 1 (total 70,000 images). Table \ref{baselinesetting} shows the result acquired by the hyper-parameter search as well as other default hyper-parameters.
\begin{table}[h!]
\centering
 \begin{tabular}{||c c c||} 
 \hline
 Optimizer name & Learning rate & Hyper-parameters \\ [0.5ex] 
 \hline
 \hline
 SGD & 0.1 & momentum=0\\ 
 \hline
 Adadelta & 1.0 & rho=0.9\\ 
 \hline
 Adagrad & 0.1 & \\ 
 \hline
 Adam & 0.001 & betas=(0.9, 0.999)\\ 
 \hline
\end{tabular}
\caption{Four different optimizers and their hyper-parameters. Figure \ref{baselinemodelsparandselec} shows the test accuracy plot of four different optimizers.}
\label{baselinesetting}
\end{table}
\newline
To train the MNIST data set, the standard feed-forward model was used. The baseline model has the following structure: an input layer with 28 x 28 = 784 neurons, a single hidden layer with 256 neurons, and an output layer with 10 neurons. Sigmoid activation function\footnote{Hoyer's sparsity and CCMAS selectivity assume that the activation value is positive. It is possible to obtain the rate that bounds between 0 and 1 by using the sigmoid activation function. Unlike the sigmoid, other activation functions such as tanh cannot be used directly to measure the sparsity and selectivity. They would need extra manipulations in the calculation. For example, the absolute value should be applied to tanh activation. Relu activation function could be used instead of the sigmoid activation function. It should be noted that the interpretation of sparsity and selectivity will be varied by applying different activation functions.} is used.
The standard feed-forward network has the following form: A softmax layer is applied to the output layer, cross-entropy loss is applied to the loss function, and the standard back-propagation algorithm updates parameters \cite{backprop}. 
\section{Sparsity and Selectivity graphs}
\begin{figure}[h!]
  \centering
  \includegraphics[width=1.0\textwidth]{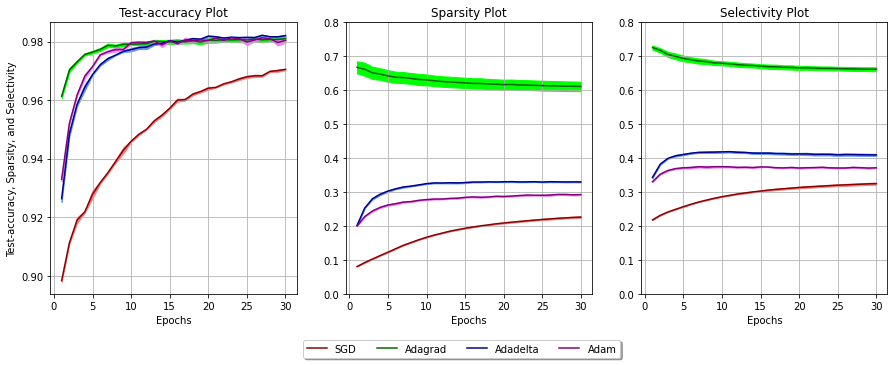}
  \caption{Test accuracy, sparsity, and selectivity. Data represented as mean and standard error (n=3)}
  \label{baselinemodelsparandselec}
\end{figure}
Figure \ref{baselinemodelsparandselec} shows the sparsity and selectivity of each epoch acquired by using the optimizers.
Firstly, it has been observed that they had similar trends.
For SGD, Adadelta and Adam, they tend to increase as epoch progresses. 
It has been observed that the test accuracy, sparsity, and selectivity arises simultaneously.
On the other hand, in Adagrad, the sparsity and selectivity tend to decrease as the epoch progresses. 
Adagrad shows a tendency of the increasing test accuracy as they decrease.
Secondly, it has been found that their values become more stable as the epoch progresses.

\section{Training vs Testing}
\begin{figure}[h!]
  \centering
  \includegraphics[width=0.7\textwidth]{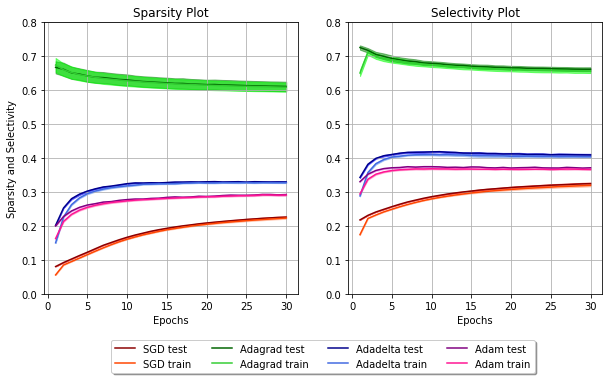}
  \caption{Sparsity and selectivity of the training and testing. Data represented as mean and standard error (n=3)}
  \label{trainingtesting}
\end{figure}

Sparsity and selectivity were measured on the testing set, although they could be measured on both the training and testing sets. 
According to Sutskever et al. \cite{sutskever2013importance}, CCMAS selectivity was calculated on the testing set; thus, sparsity was calculated on the testing set for consistency.
There may be differences in the sparsity and selectivity values in the training and testing; hence, this is analyzed by Figure \ref{trainingtesting}.
The training and testing result were acquired by the baseline model.
\begin{figure}[h!]
  \centering
  \includegraphics[width=0.7\textwidth]{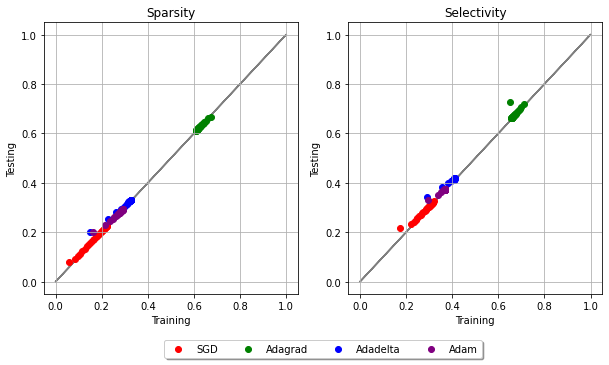}
  \caption{Comparing sparsity and selectivity values of the training and testing.}
  \label{trainingtesting2}
\end{figure}
\newline 
By looking at Figure \ref{trainingtesting}, it can be seen that the difference between the sparsity and selectivity of the training and testing decrease as the epoch progresses, eventually ending up with a very similar value. 
Figure \ref{trainingtesting2} was plotted to get a closer look at the relationship between the training and testing.
The diagonal line of the two graphs is $y=x$. 
The closer the dots are to the diagonal line, the less difference there is between the training and testing.
It can be observed that some outliers deviate from $y=x$. 
However, these occurred at the beginning of the epoch. As values reach the final epoch, it can be seen that they converge to $y=x$. 
There is little difference in the sparsity and selectivity of the training and testing.

\section{Varying the number of hidden layers} \label{hiddenlayerlabel}
\subsection{Experimental Details}
The method of calculating sparsity and selectivity introduced in the \nameref{chap:technical} chapter was based on a single hidden layer. 
However, several hidden layers are used in this experiment; thus, obtaining their metrics would be slightly different. 
When there are multiple hidden layers, they are calculated as follows:
\begin{equation} 
\begin{split}
    \textit{sparsity of N hidden layers} &= \frac{\sum_{i=1}^{N} {sparsity\_of\_HL_N}}{N}\\
    \textit{selectivity: average of N hidden layers} &= \frac{\sum_{i=1}^{N} {selectivity\_avg\_of\_HL_N}}{N}\\
    \textit{selectivity: standard deviation of N hidden layers} &= \frac{\sum_{i=1}^{N} {selectivity\_std\_of\_HL_N}}{N}
\end{split}
\end{equation}
$HL_N$ denotes $N$-th Hidden Layer (HL). That is, sparsity, an average of selectivity, and a standard deviation of selectivity of each layer are summed and divided by the total number of hidden layers.
\newline 
\newline 
5 hidden layers were used at maximum for Adadelta and Adam and 4 hidden layers were used at most for SGD and Adagrad. For each number of hidden layers, test accuracy, sparsity, and selectivity were investigated. The number of hidden layers could not be increased more because SGD and Adagrad could not find an appropriate local optima result in very low test accuracy when there are 5 hidden layers. The network structure for each optimizer is described in Table \ref{hiddenlayertable}. 
\begin{table}[h!]
\centering
 \begin{tabular}{||c | c | c | c||} 
 \hline
 HL size & Network Structure & Batch size and hyper-parameters\\ [0.5ex] 
 \hline
 \hline
 HL 1 & I $\rightarrow$ HL $\rightarrow$ O & $\cdot$ \\ 
 \hline
 HL 2 & I $\rightarrow$ HL $\rightarrow$ HL $\rightarrow$ O & $\cdot$ \\ 
 \hline
 HL 3 & I $\rightarrow$ HL $\rightarrow$ HL $\rightarrow$ HL $\rightarrow$ & $\cdot$ \\ 
 \hline
 HL 4 & I $\rightarrow$ HL $\rightarrow$ HL $\rightarrow$ HL $\rightarrow$ HL $\rightarrow$ O & $\cdot$ \\ 
 \hline
 HL 5 & I $\rightarrow$ HL $\rightarrow$ HL $\rightarrow$ HL $\rightarrow$ HL $\rightarrow$ HL $\rightarrow$ O & $\cdot$ \\ 
 \hline
\end{tabular}
\caption{I = Input = 756 neurons, HL = Hidden Layer = 256 neurons, O = Output = 10 neurons. $\cdot$ means \textit{batch size} = 50 and it uses the baseline hyper-parameters presented in Table \ref{baselinesetting}.}
\label{hiddenlayertable}
\end{table}
\subsection{Overall Results}
\begin{figure}[h!]
  \centering
  \includegraphics[width=1.0\textwidth]{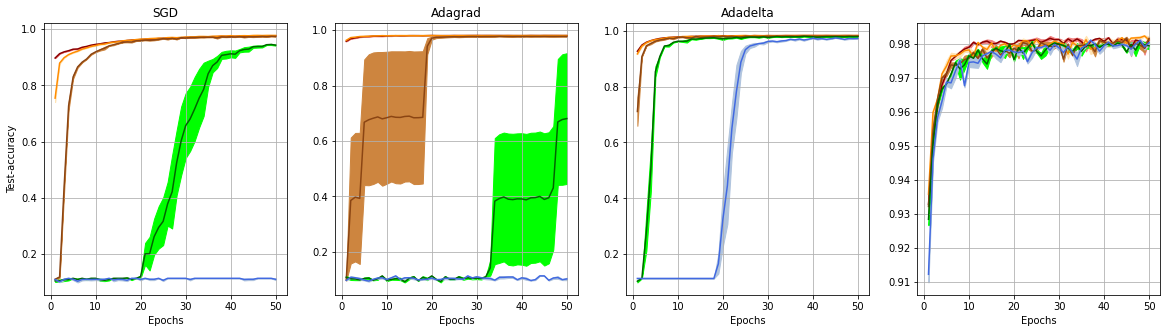}
  \includegraphics[width=1.0\textwidth]{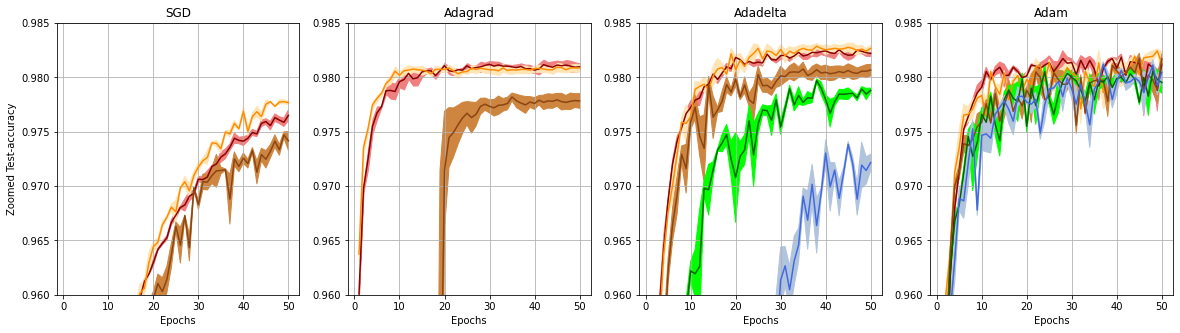}
  \includegraphics[width=1.0\textwidth]{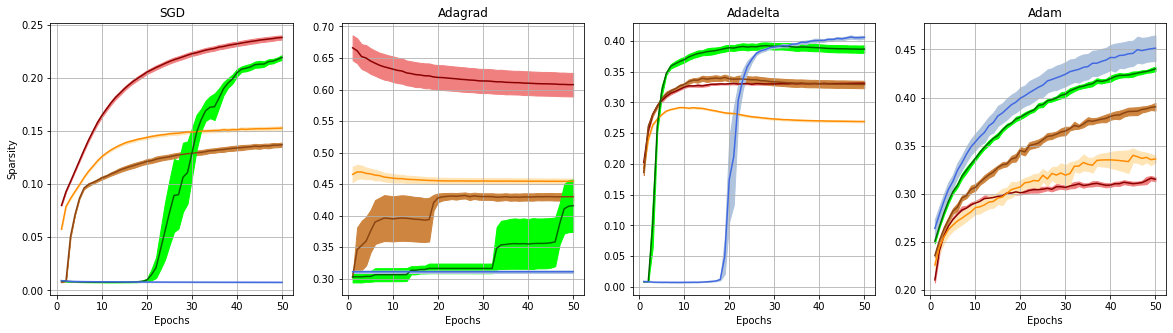}
  \includegraphics[width=1.0\textwidth]{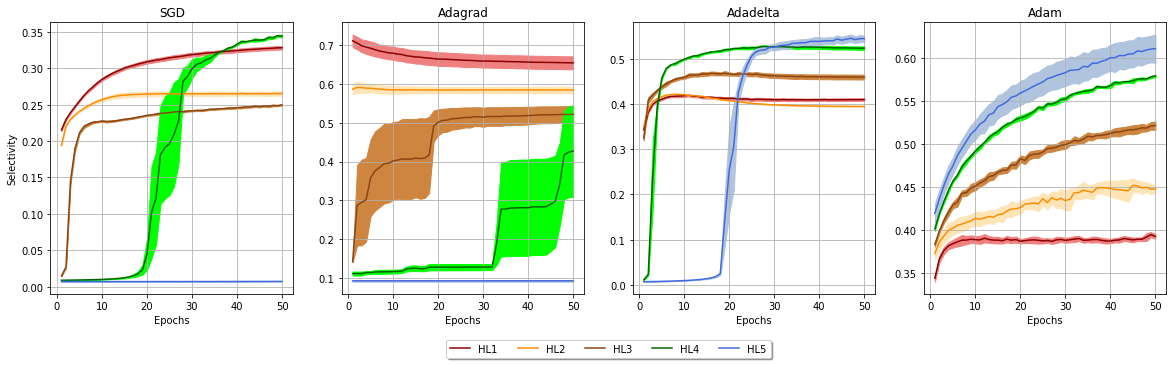}
  \caption{Test Accuracy, sparsity, and selectivity of 4 optimizers with multiple hidden layers. Data represented as mean and standard error (n=3)}
  \label{alloptimhlSM}
\end{figure}

Figure \ref{alloptimhlSM} (See SM) shows the test accuracy, sparsity, and selectivity of four optimizers by varying the number of hidden layers. 
The figure shows that the sparsity and selectivity of Adagrad HL1 and HL2 are different from the three other sparsity and selectivity trends. 
Their values tend to decrease as the epoch progresses. 
The other optimizers' values tend to increase as epochs progress. 
In this section, a total of four topics will be discussed by extracting essential features shown in Figure \ref{alloptimhlSM}. 
\begin{enumerate}
\item The relationship between sparsity and selectivity
\item The relationship between test accuracy, sparsity, and selectivity. 
\item The test accuracy, sparsity, and selectivity with different number of hidden layers.
\item The test accuracy, sparsity, and selectivity of each hidden layer.
\end{enumerate}
\subsection{Sparsity vs Selectivity 1}
\label{sparvsselecsection1}
\begin{figure}[h!]
  \centering
  \includegraphics[width=1.0\textwidth]{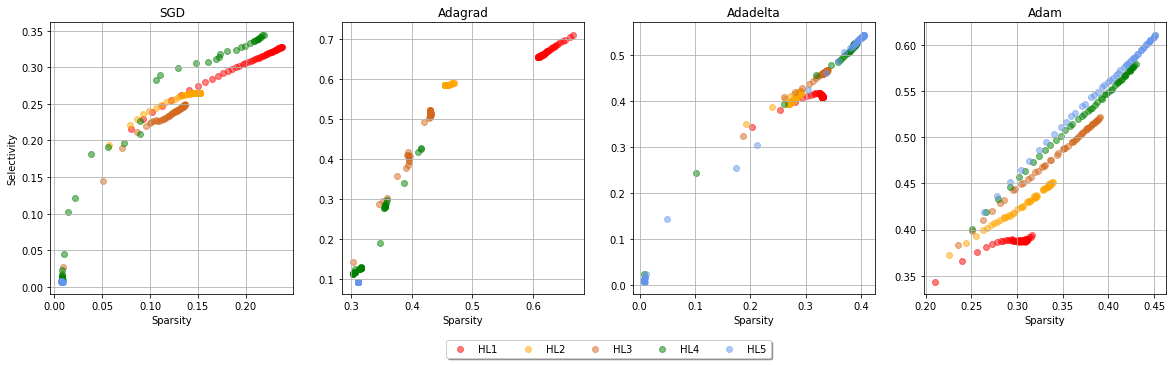}
  \caption{Comparing sparsity and selectivity. It should be noted that the scatter graph does not take epoch into account; therefore, the increasing sparsity and selectivity does not always mean that the epoch is progressing.}
  \label{sparvsselec}
\end{figure}

By looking at Figure \ref{alloptimhlSM}, it can be observed that the sparsity and selectivity tend to arise at the same time. 
This result supports the idea: sparsity encourages feature selective response \cite{sharkey1995analysis, murre2014learning}. Figure \ref{sparvsselec} is plotted to investigate the relationship in more detail. 
\newline 
It turned out that the sparsity did not always encourage feature selective response.
Although the sparsity and selectivity commonly tend to arise simultaneously, there are exceptions.
If sparsity always enables selectivity, the scattered dots should always move towards the top right corner (or bottom left corner in case of Adagrad HL1 and HL2). 
This phenomenon can be clearly observed in SGD HL2, Adadelta HL1, and Adam HL1. 
SGD HL2 shows that as the sparsity increases, the selectivity remains relatively constant.
Adadelta HL1 show that as the sparsity increases, the selectivity decreases.
Adam HL1 shows that as the sparsity increases, the selectivity remains relatively constant.
\subsubsection{Conclusion}
Although the sparsity and selectivity mostly arise simultaneously, sometimes they do not. Such notion supports the idea that sparsity does not always yield selectivity \cite{spar_selec_does_not_related, willmore2001characterizing}. 
\subsection{Accuracy vs Sparsity and Selectivity 1}
\label{accuracyvspsarselec1}
\begin{figure}[h!]
  \centering
  \includegraphics[width=1.0\textwidth]{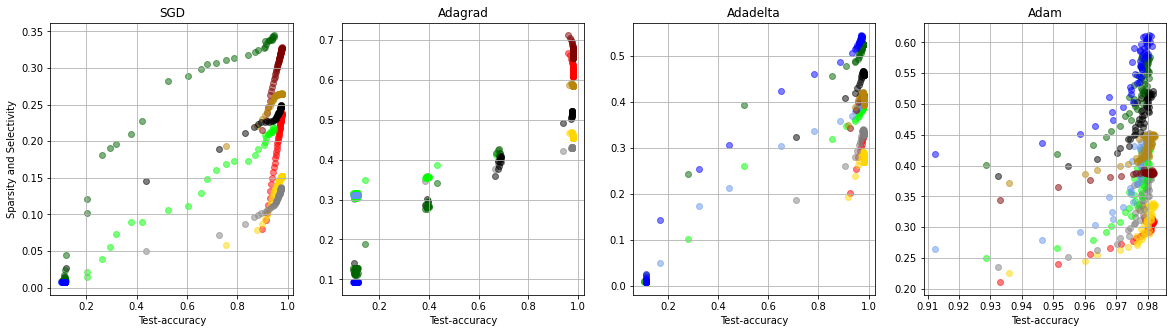}
  \includegraphics[width=1.0\textwidth]{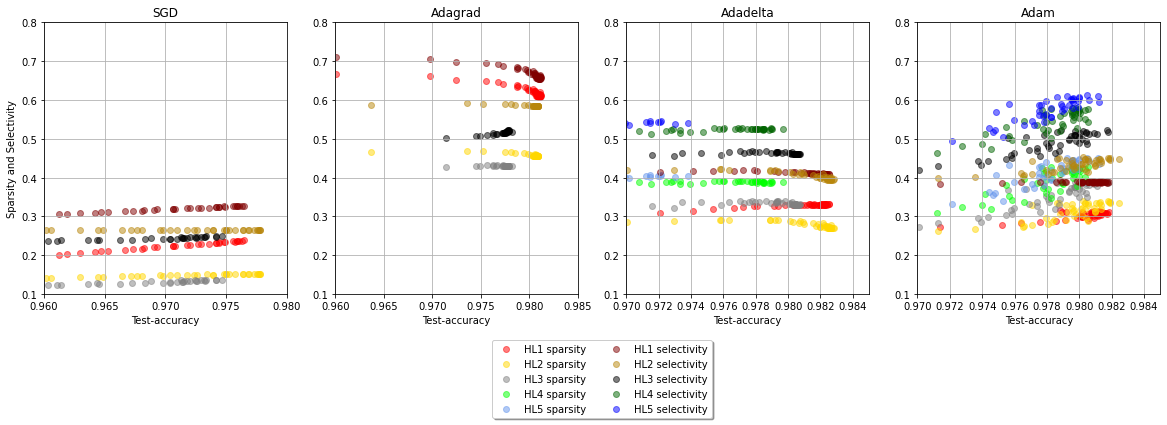}
  \caption{Top: Test accuracy vs sparsity and selectivity. Bottom: Zoomed in test accuracy vs sparsity and selectivity. It only shows the mean of the sparsity and selectivity average. The standard error is omitted. It should be noted that the scatter graph does not take the epoch into account; therefore, the increasing test accuracy does not always mean that the epoch has progressed.}
\label{accsparselec}
\end{figure}

When observing the test accuracy, sparsity, and selectivity of Figure \ref{alloptimhlSM}, it is likely that they are related. They increase or decrease simultaneously. 
This can be explained by the following:
Assume that weights (parameters) can be tracked by $A$ and $A$ reside in the parameter space $S$. $A$ moves around $S$ during learning and adjusts the weights. At the beginning of learning, the local basin has not yet been found; thus, it wanders $S$ without specific direction. 
It implies that most weights are used to find a local basin.
This encourages a less sparsity and selectivity.
After finding an appropriate local basin, only certain weights are likely to be updated since the update orients toward the direction of local minimum. The closer to the local minimum, the higher the probability that only certain weights will be updated. This encourages increasing sparsity and selectivity as well as increasing accuracy.
\newline 
\newline 
To investigate this in more detail, the scatter graph of test accuracy vs sparsity and selectivity was plotted in Figure \ref{accsparselec}.
It was found that the tendency of the test accuracy, sparsity, and selectivity was similar; but, not always. In other words, it cannot be guaranteed that the value of the sparsity and selectivity always increases when the test accuracy increases. 
Similarly, it cannot be guaranteed that the value of the test accuracy always increases when they increase. 
For example, they fluctuate when the test accuracy is constant (e.g. See when Adam's test accuracy is 0.978).
\newline 
\newline 
Another thing to note here is the relationship between the test accuracy, sparsity, and selectivity in Adadelta (See Adadelta - the bottom figures of Figure \ref{accsparselec}).
It has been already observed that Adagrad HL1 and HL2 decrease its sparsity and selectivity when the test accuracy increases. 
Adadelta HL2 and HL3 shows a similar tendency.
Adadelta HL2 and HL3 decrease them when the test accuracy increases. This might indicate that in order to increase the test accuracy further, the sparsity and selectivity should be decreased. This conforms with the argument from Leavitt et al. \cite{leavitt2020selectivity} where they noted test accuracy has been improved by reducing selectivity. I argue that test accuracy can be improved by reducing sparsity as well.
\subsubsection{Conclusion}
The test accuracy, sparsity, and selectivity tend to arise simultaneously in general, which means they seem to be related; however, this is not always the case. This experiment additionally supports Leavitt et al.'s argument \cite{leavitt2020selectivity}: Test accuracy is unaffected by reducing class selectivity (See Adam in Figure \ref{accsparselec}). In addition to their argument, it can be derived that test accuracy is sometimes unaffected by reducing sparsity. Furthermore, based on the analysis, it can be concluded that the test accuracy cannot precisely predict the sparsity and selectivity and vice versa.
\subsection{Accuracy vs Sparsity and Selectivity 2}
\label{accuracyvspsarselec2}
By observing only the last epoch of each experiment (HL1, HL2, HL3, HL4 and HL5) in Figure \ref{alloptimhlSM}, it seems like the test accuracy decreases when the sparsity and selectivity increase. For example in Adam, the test accuracy decreases along with the increment of hidden layers; however, the sparsity and selectivity increase. Similar phenomena could be observed in Adadelta HL4 and HL5. To investigate this in more detail, Table \ref{extracted} is shown.
\begin{table}[h!]
\centering
 \begin{tabular}{||c | c | c | c||} 
 \hline
 Optimizer name & Accuracy & Sparsity & Selectivity avg\\ [0.5ex] 
 \hline
 \hline
 SGD HL 1 & 0.9765 & 0.238 & 0.3279 \\
 \hline
 SGD HL 2 & 0.9776 & 0.153 & 0.2652\\
 \hline
 SGD HL 3 & 0.9742 & 0.137 & 0.2495\\
 \hline
 SGD HL 4 & 0.943  & 0.219 & 0.3439\\
 \hline
 \hline
 Adagrad HL 1 & 0.981 & 0.6077 & 0.6542 \\
 \hline
 Adagrad HL 2 & 0.981 & 0.4544 & 0.584\\
 \hline
 Adagrad HL 3 & 0.978 & 0.430 & 0.512\\
 \hline
 Adagrad HL 4 & 0.6812  & 0.415 & 0.426\\
 \hline
 \hline
 Adadelta HL 1 & 0.982 & 0.331 & 0.410 \\
 \hline
 Adadelta HL 2 & 0.983 & 0.269 & 0.395\\
 \hline
 Adadelta HL 3 & 0.981 & 0.329 & 0.460\\
 \hline
 Adadelta HL 4 & 0.979  & 0.387 & 0.524\\
 \hline
 Adadelta HL 5 & 0.972  & 0.406 & 0.544\\
 \hline
 \hline
 Adam HL 1 & 0.981 & 0.315 & 0.392 \\
 \hline
 Adam HL 2 & 0.981 & 0.336 & 0.448\\
 \hline
 Adam HL 3 & 0.981 & 0.390 & 0.522\\
 \hline
 Adam HL 4 & 0.980  & 0.430 & 0.580\\
 \hline
 Adam HL 5 & 0.980  & 0.451 & 0.611\\
 \hline
\end{tabular}
\caption{Test accuracy, sparsity, and selectivity of optimizers at the last epoch. HL5 for SGD and Adagrad is omitted due to low accuracy. The table is used instead of the graph because it could not represent minute differences between test accuracy, sparsity, and selectivity well.}
\label{extracted}
\end{table}
These phenomena could be observed in SGD HL3 to HL4, Adadelta HL2 to HL5, and Adam HL1 to HL5. It can be observed that the sparsity and selectivity tend to increase when the test accuracy decreases. 
In addition, it can be observed that they tend to decrease when the test accuracy increases in SGD HL1 to HL2 and Adadelta HL1 to HL2. 
\subsubsection{Conclusion}
These tendencies are consistent with the argument from Leavitt et al. \cite{leavitt2020selectivity}: Increased class selectivity is considered harmful, and test accuracy is improved by reducing selectivity. Through this experiment, I argue that increased sparsity is considered harmful and test accuracy can be improved by reducing sparsity. 
It should be noted that similar findings were discussed in Gale et al. \cite{gale2019state} where high sparsity shows low test accuracy.
\subsection{Varying the number of hidden layers}
In this subsection, this paper will examine test accuracy, sparsity, and selectivity of each optimizer by varying the number of hidden layers. Since the human brain resembles a multiple hidden layers structure than a single hidden layer structure, it has been expected that sparsity and selectivity would increase by the addition of hidden layers. 
\begin{figure}[h!]
  \centering
  \includegraphics[width=1.0\textwidth]{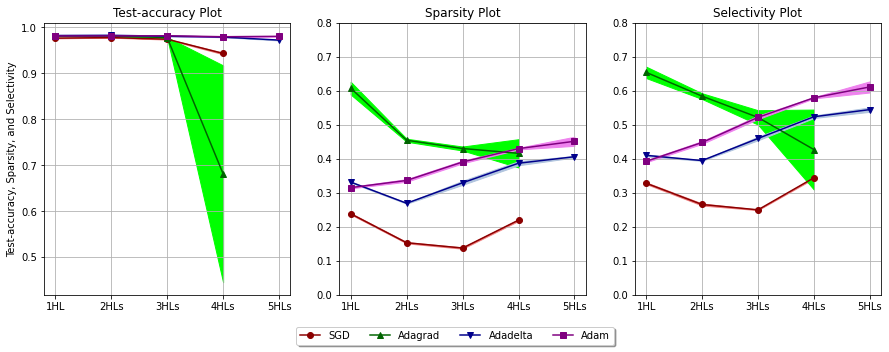}
  \caption{Test accuracy, sparsity, and selectivity when varying a number of hidden layers. This graph is the plotted version of Table \ref{extracted}. Data represented as mean and standard error (n=3)}
  \label{ssvsnumberofhiddenlayer}
\end{figure}
However, although Adadelta and Adam show the expected likelihood, SGD and Adagrad do not. Adadelta and Adam gradually increase their sparsity and selectivity as the hidden layer is added to the network. Adagrad decreases its sparsity and selectivity as the hidden layer is added to the network. 
There is no clear tendency in SGD; they fluctuate with an increasing number of the hidden layer.
\subsubsection{Conclusion}
There is no clear relationship between the sparsity and selectivity and the increasing number of hidden layers. Different optimizer shows different tendency. Adadelta and Adam could be considered to increase them. 
It should be noted that the cost of decrements in the test accuracy should be taken into account to acquire high sparsity and selectivity with many hidden layers.
\subsubsection{Limitations} \label{hiddenlayerlimitations}
Firstly, it was difficult to increase more hidden layers in SGD and Adagrad; because the performance was very bad (very low test accuracy, sparsity, and selectivity. See Figure \ref{alloptimhlSM}) when there were more than 4 hidden layers with 256 neurons each. In order to circumvent this problem, it may be possible to reduce the number of neurons in the hidden layer and then increase the number of the hidden layers. 
For example, 5 hidden layers with 64 neurons each may be possible to find a suitable local optimum. More hidden layers can be implemented into the network to investigate sparsity and selectivity if this is possible. As a result, it would be possible to acquire a clear tendency with an increasing number of hidden layers.
\newline 
\newline 
Secondly, in the process of obtaining the sparsity and selectivity of multiple hidden layers, I summed all sparsity and selectivity values and averaged by the number of hidden layers; thus, they may not have been accurately measured. 
I propose possible sparsity and selectivity metrics for the multiple hidden layers network.
\begin{equation} 
\begin{split}
    \textit{activation of $HL_i$} &= {\sum_{n=1}^{N} \underline{\text{$activation\_val_n$ in $HL_i$}}}\\
    \textit{$activation\_ratio_k$} &= {\frac{\textit{activation of $HL_k$}}{\sum_{i=1}^{M} \textit{activation of $HL_i$}}}\\
    \textit{Sparsity} &= {\sum_{k=1}^{M} \text{$activation\_ratio_k$} \times \underline{\textit{Sparsity $HL_k$}}}\\
    \textit{Selectivity} &= {\sum_{k=1}^{M} \text{$activation\_ratio_k$} \times \underline{\textit{Selectivity $HL_k$}}}
\end{split}
\end{equation}
where \underline{\text{$activation\_val_n$ in $HL_i$}} denotes an activation value for $n-th$ neuron in the $i-th$ hidden layer (HL). $N$ denotes the number of neurons in $HL_i$. M represents the number of hidden layer in neural network. \underline{\textit{Sparsity $HL_k$}} and \underline{\textit{Selectivity $HL_k$}} denotes sparsity and selectivity values of $HL_k$. Sparsity and selectivity of each hidden layer are multiplied by a different ratio and the ratio indicates how active that the particular hidden layer is compared to other hidden layers. 
\subsection{Sparsity and Selectivity of each hidden layer}
\begin{figure}[h!]
  \centering
  \includegraphics[width=0.7\textwidth]{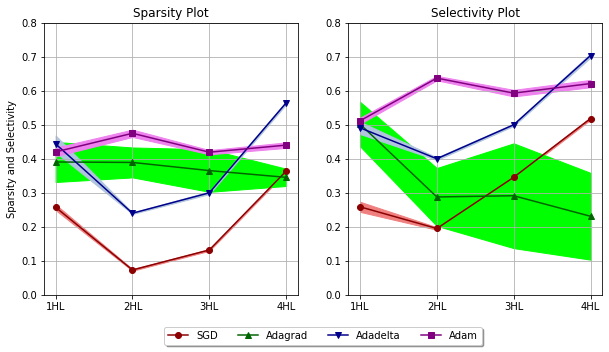}
  \caption{Sparsity and selectivity of each hidden layer extracted at the last epoch. There are four hidden layers. Data represented as mean and standard error (n=3).}
  \label{hl_each_extracted}
\end{figure}
To investigate the relationship between \textit{each} hidden layer, sparsity, and selectivity, Figure \ref{hl_each_extracted} is plotted. 
In SGD and Adadelta, it can be observed that the sparsity and selectivity of HL1 are quite high. 
This may arise because edge\footnote{Top left, top right, bottom left, bottom right} pixels of the image in MNIST data is always 0 (black). 
Weights for those connections to HL1 are not going to learn anything\footnote{HL1 will ignore those 0 (black) inputs}, resulting high sparsity and selectivity. HL2 shows low sparsity and selectivity compared to HL1 since HL1 is fully connected with HL2. 
The above explanation sounds plausible; however, it cannot explain why a similar phenomenon does not occur in Adagrad and Adam. It is difficult to explain why these phenomena occur. Overall, there is no clear tendency in the sparsity and selectivity of \textit{each} hidden layer. 
\subsubsection{Conclusion}
Even though there is no clear relationship between them, it may be possible to conclude that the sparsity and selectivity increase when the hidden layer deepens (HL2 to HL4) in SGD and Adadelta, considering that the sparsity and selectivity of HL1 are quite high because of the sparsity of the image itself. They decrease when the hidden layer deepens in Adagrad. There is no clear tendency in Adam. This topic should be further researched in order to be generalized.


\section{Varying the Hyper-parameters}
\label{varyinghyperparamlabel}
\subsection{Experimental details}
In this section, this paper will compare test accuracy, sparsity, and selectivity of each optimizer while changing hyper-parameters. Also, if sparsity and selectivity increase, the paper will try to analyse which characteristics caused them to arise. 
The learning rate and weight decay are common to all optimizers. Additionally, SGD has a $\gamma$ hyper-parameter (see subsection \ref{sgdmomentum}), Adadelta has a $\rho$ hyper-parameter (see subsection \ref{Adadeltarho}), and Adam has $\beta_1$ and $\beta_2$ hyper-parameters (see subsection \ref{adambetas}). The number of epochs is set to 30. 
The values of the sparsity and selectivity were extracted at the final epoch.
For detailed experimental details, refer to the following table.
\begin{table}[h!]
\centering
 \begin{tabular}{||c | c | c | c||} 
 \hline
 Optimizer name & Experimented hyper-parameters\\ [0.5ex] 
 \hline
 \hline
 Common & weight decay = np.logspace(-5, 1, 11)\\
 \hline 
 Common & learning rate = np.logspace(-5, 3, 15)\\
 \hline 
 SGD & momentum = [0.0, 0.1, 0.2, 0.3, 0.4, 0.5, 0.6, 0.7, 0.8, 0.9, 0.999]\\
 \hline
 Adagrad & $\cdot$\\
 \hline 
 Adadelta & rho = [0.0, 0.1, 0.2, 0.3, 0.4, 0.5, 0.6, 0.7, 0.8, 0.9, 0.999]\\
 \hline 
 Adam & betas = [0.0001, 0.1, 0.2, 0.3, 0.4, 0.5, 0.6, 0.7, 0.8, 0.9, 0.999]\\
 \hline
\end{tabular}
\caption{The range of experimented hyper-parameter. $\cdot$ indicates there is no additional varying hyper-parameter property.
np indicates numpy library in Python 3.}
\label{varyinghyperparamtable}
\end{table}
\subsection{Results}
\subsubsection{Weight decay}
\begin{figure}[h!]
  \centering
  \includegraphics[width=1.0\textwidth]{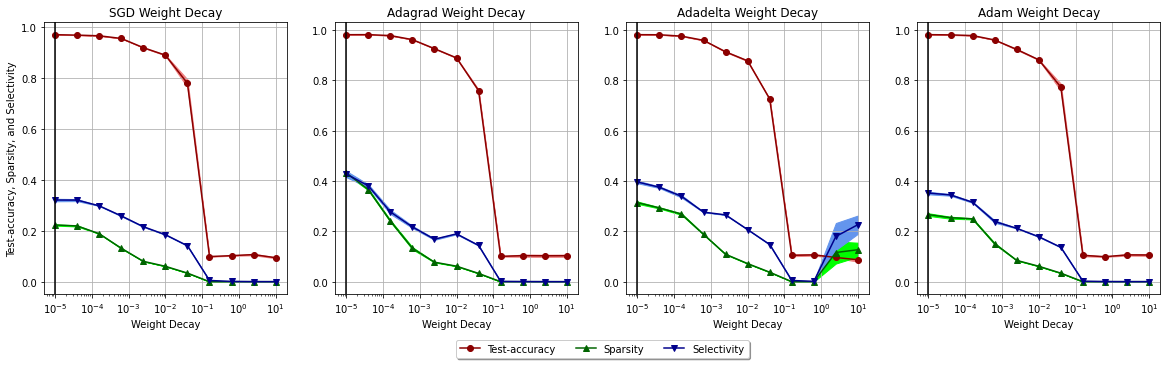}
  \caption{Test accuracy, sparsity, and selectivity by varying the weight decay hyper-parameter. The black vertical line indicates the default value of the weight decay hyper-parameter. Data represented as mean and standard error (n=3).}
  \label{wdhyper}
\end{figure}
By looking at Figure \ref{wdhyper}, it was found that adding the weight decay for all four optimizers had a negative effect on the sparsity and selectivity. It seemed like they tend to decrease because the test accuracy decreases. It has been found that the test accuracy, sparsity, and selectivity tend to increase or decrease simultaneously in subsection \ref{accuracyvspsarselec1}.
\subsubsection{Learning rate}
\begin{figure}[h!]
  \centering
  \includegraphics[width=1.0\textwidth]{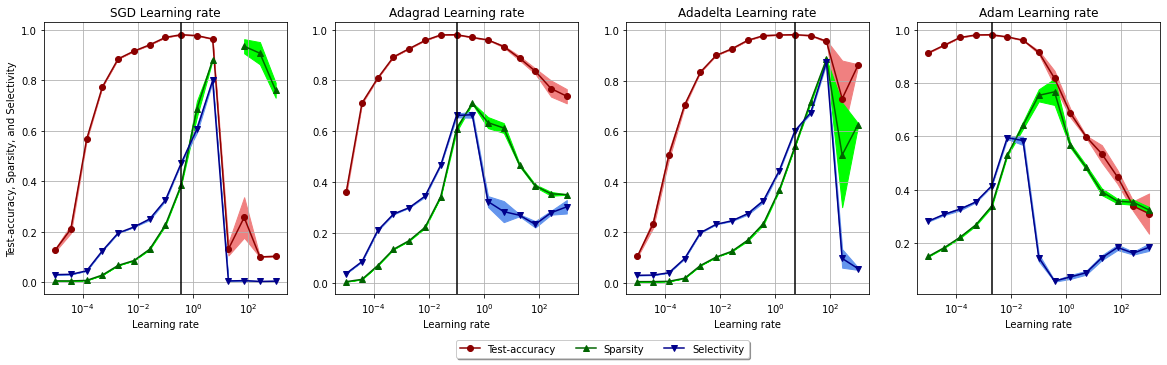}
  \caption{Test accuracy, sparsity, and selectivity by varying the learning rate hyper-parameter. There is one missing value in SGD Learning rate plot. It encountered $nan$ value. The black straight line indicates an optimal learning rate for the highest test accuracy. Data represented as mean and standard error (n=3).}
  \label{wdlr}
\end{figure}
\noindent Figure \ref{wdlr} shows the relationship between the learning rate, test accuracy, sparsity, and selectivity. 
One of the most important things to note here is that the highest sparsity and selectivity are acquired when the learning rate is slightly higher than the optimal learning rate. 
It is common to aim for an optimal learning rate based on test accuracy without considering sparsity and selectivity. 
It can be induced that the conventional method of searching for an optimal learning rate based on test accuracy is far away from human's continual learning. 
This may have arisen because a high learning rate fluctuates more than a small learning rate when training.
It may be more reasonable to take sparsity and selectivity into account as well as test accuracy when choosing an optimal learning rate.
As we still have a limited understanding of a large learning rate, this result may provide insights to it.
\newline 
\newline
A few more interesting phenomenon can be observed.
Firstly, the commonly observable feature of all four graphs is that the test accuracy, sparsity, and selectivity increase simultaneously until the test accuracy reaches the highest value. 
Secondly, it can be seen that there is a trade-off between the test accuracy, sparsity, and selectivity. In SGD graph, the sparsity and selectivity do not have the highest value around $learning\_rate=0.1$ where the test accuracy is the highest; instead, their highest values are located in $learning\_rate=[1, 10]$. 
Similar phenomena could be found on other graphs. This phenomenon supports the conclusion from subsection \ref{accuracyvspsarselec2}: Increased sparsity and selectivity considered harmful to test accuracy. 
Thirdly, it can be seen that the sparsity and selectivity do not occur at the same time. This phenomena only occurs after reaching the peak of accuracy. For example, in SGD, the sparsity after $learning\_rate=10^2$ is very high, although the selectivity stays 0. In Adam, the sparsity starts to decrease around $learning\_rate=1$, but the selectivity begins to fall around $learning\_rate=10^{-2}$. This phenomenon supports the conclusion from subsection \ref{sparvsselecsection1}: Sparsity and selectivity may not arise simultaneously. 
\subsubsection{Other hyper-parameters}
\label{otherhplabel}
\begin{figure}[h!]
  \centering
  \includegraphics[width=1.0\textwidth]{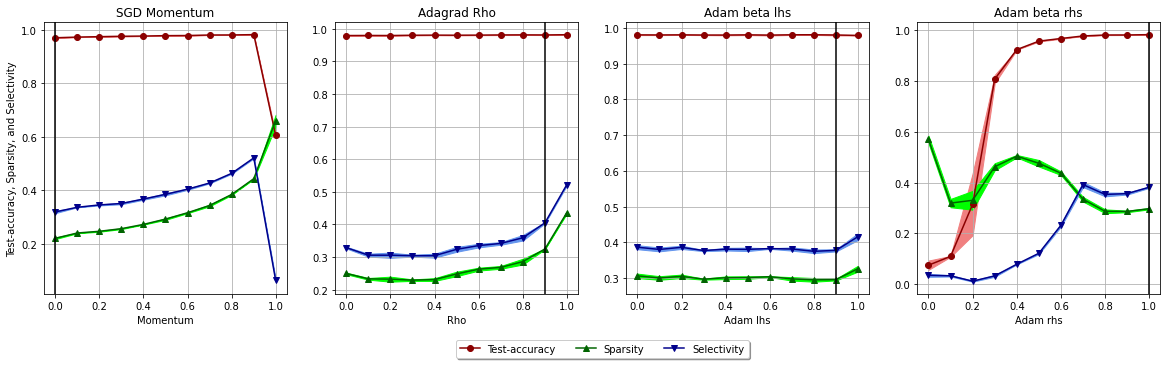}
  \caption{Test accuracy, sparsity, and selectivity by varying other hyper-parameters. The black vertical line indicates the default value of each hyper-parameter. $lhs=\beta_1$ and $rhs=\beta_2$. Data represented as mean and standard error (n=3).}
  \label{hyperhyper}
\end{figure}
\noindent By looking at Figure \ref{hyperhyper}, in SGD momentum and Adadelta rho graphs, it can be clearly seen that $\gamma$ value and $\rho$ value affect the sparsity and selectivity.
In the case of $\gamma$, they tend to increase as $\gamma$ increases.
High fluctuations when training increased the sparsity and selectivity (See equation \ref{sgdmomentumeq}).
However, if $\gamma$ is too large, their values drop. 
In the case of $\rho$, they tend to increase as $\rho$ increases specifically after 0.4.
In the case of Adam, their tendency cannot be clearly seen. They slightly increased in the range from 0.9 to 1.0\footnote{It is more reasonable to observe between [0.9, 1.0] in Adam; because the algorithm is designed to have the high value of $\beta_1=0.9$ and $\beta_2=0.999$.}.
These facts clearly show a relationship between the degree of taking past gradients, sparsity, and selectivity. 
By observing them in Adam beta rhs graph, it can be seen that they do not tend to increase or decrease together. The sparsity decreases in range from 0.4 to 0.7; whereas, the selectivity increases in a range from 0.4 to 0.7. This conforms with the conclusion from subsection \ref{sparvsselecsection1}.
\subsection{Conclusion}
\begin{enumerate}
    \item Weight Decay: The weight decay does not help the sparsity and selectivity to arise. 
    \item Learning rate: The optimal learning rate does not guarantee the highest sparsity and selectivity value; rather, they aroused when the learning rate was slightly higher than the optimal learning rate.
    \item Other hyper-parameters: It is likely that the more past gradient is taken into account when updating parameters, the higher value of the sparsity and selectivity is gained. However, looking at SGD momentum graph, if $\gamma$ is too large, then the test accuracy and the selectivity will drop.
\end{enumerate}

\section{Varying the Batch Size}
\label{varyingbatchsize}
\subsection{Experimental details}
Despite the benefits of varying a batch size or equivalently data parallelism, understanding its effects in a neural network is still unclear.
In this section, this paper will compare test accuracy, sparsity, and selectivity values by varying the batch size. If there is any change in these, I will analyse which property caused sparsity and selectivity to be increased. The attempted batch sizes are 1, 5, 10, 50, 200, 500, 1000. For detailed experimental details, refer to the following table.
\begin{table}[h!]
\centering
 \begin{tabular}{||c | c | c | c||} 
 \hline
 Optimizer name & Batch sizes & hyper-parameters\\ [0.5ex] 
 \hline
 \hline
 All & batch size = [1, 5, 10, 50, 200, 500, 1000] & $\cdot$\\
 \hline 
\end{tabular}
\caption{Varying the batch size for four optimizers. $\cdot$ means the baseline hyper-parameters are used. Epoch is set to 30.}
\label{varyingbatchsizetable}
\end{table}
\subsection{Results}
\begin{figure}[h!]
  \centering
  \includegraphics[width=1.0\textwidth]{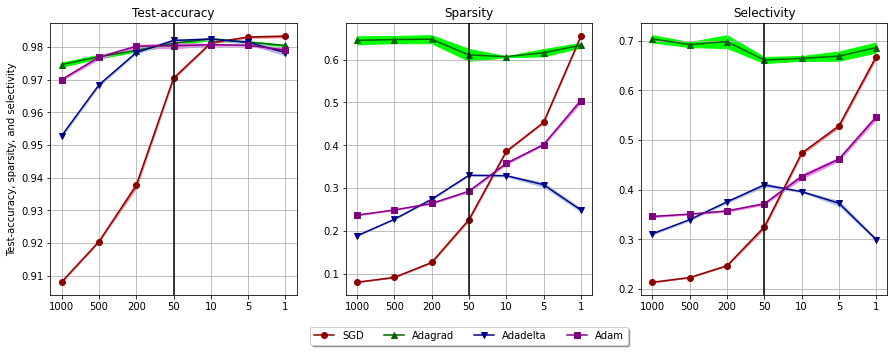}
  \caption{Test accuracy, sparsity, and selectivity by varying the batch size. The values of test accuracy, sparsity, and selectivity were extracted at the last epoch. The black vertical line indicates the baseline model. Data represented as mean and standard error (n=3).}
  \label{varyingbactchsizefigure}
\end{figure}
\noindent Looking at Figure \ref{varyingbactchsizefigure}, a few phenomena could be observed.
Firstly, the test accuracy of SGD tends to increase with a smaller batch size. 
It outperforms other optimizers when $bs=1$ (\textit{batch size} = 1). 
It performs better or similar to other adaptive optimizers in the sparsity and selectivity. 
Secondly, the sparsity and selectivity increased significantly even though the accuracy hardly changed in SGD and Adam, specifically when \textit{batch size} = 1, 5. 
Thirdly, it looks like there is an optimal batch size for the highest sparsity and selectivity depending on each optimizer. 
They increased with a smaller batch size in SGD and Adam. They seem to be unrelated to the batch size in Adagrad. They have the highest value when the batch size is 50 in Adadelta. 
\newline
\newline 
This paper mainly focuses on sparsity and selectivity. Nevertheless, it will explain about test accuracy regarding the batch size as there seems to be a close relation between the batch size, test accuracy, sparsity, and selectivity. Also, the explanation might provide useful insights for future work.
The followings could describe the explanation of the phenomenon where the SGD's test accuracy of the smaller batch size surpasses the larger batch size and outperforms other adaptive optimizers.
According to section \ref{batchsizerelationship}, in SGD, if the batch size is small, the latest gradient can be used for updating the parameter. 
On the contrary, if the batch size is large, the latest gradient cannot be used. 
For this reason, smaller batch tends to be more generalized than larger batch \cite{masters2018revisiting}. 
Similarly, Keskar et al. \cite{keskar2016large} support this idea: small batch works well for generalization in SGD. 
In addition, Xie et al. \cite{xie2020diffusion} mathematically proved that training with a large batch is not efficient for searching flat minimum where it is more likely to acquire generalized performance in SGD.
The test accuracy plot of Figure \ref{varyingbactchsizefigure} supports the analysis of aforementioned researches\footnote{It does not mean that a larger batch always performs worse than the smaller batch. According to Shallue et al. \cite{shallue2018measuring}, there is no evidence to say the larger batch degrades the generalization performance.} \cite{masters2018revisiting, keskar2016large, xie2020diffusion}. 
Although these can prove why SGD performs better when the batch size is small, it is not enough to prove why SGD works better than adaptive methods. 
Numerous studies have empirically proved that adaptive methods may impair generalization ability leading to low testing accuracy \cite{keskar2017improving, chen2018closing}. 
It has been evaluated that SGD sometimes shows better generalization ability than Adagrad, RMSProp, and Adam \cite{wilson2017marginal}. 
Likewise, Meng et al. \cite{meng2020dynamic} showed that SGD could efficiently deviate from a sharp minimum and converge to a flat minimum.
Specifically, Zhou et al. \cite{zhou2020towards} explain why SGD has a better generalization ability than Adam in terms of local convergence behaviour\footnote{Zhou et al. \cite{zhou2020towards} stated that SGD is better than Adam in terms of the escaping time of Levy-driven stochastic differential equations from a local basin. This is because the exponential moving average of Adam leads to less gradient noise than SGD; thus, SGD can escape from a sharp minimum than Adam and more likely to converge to a flat minimum.}. 
\newline 
\newline
The aforementioned researches above clearly show that there is a relationship between the batch size and test accuracy. 
I tried to relate these researches to the sparsity and selectivity.
However, it was hard to reason how the batch size is related to them.
It is known that a smaller batch size tends to fluctuate more than a larger batch size.
Possibly, sparsity and selectivity are related to fluctuation for searching local minimum.
As the shape of local optimum (sharp, flat, or asymmetric valley\footnote{Asymmetric valleys is a word that aroused from He et al. \cite{he2019asymmetric} which refers to a local minimum that the loss increases abruptly on the one side, increases mildly on the other side. They mentioned about this word because it was observed that there are many asymmetric valleys which are quite different from just flat or sharp minimum in loss function.}) is closely related to generalization performance, can sparsity and selectivity be explained by the shape of local basin? 
\subsection{Sparsity and Selectivity 2}
\label{sparvsselecsection2}
A few notable tendencies of the sparsity and selectivity could be observed in Figure \ref{varyingbatchaccsparselecSM} (See SM). 
Firstly, it has been examined that the tendency of the varying sparsity and selectivity across epoch changes by varying the batch size.
For example, it has been observed that the sparsity and selectivity begin with a low value and gradually increase except for Adagrad in the baseline model. 
However, it can be seen that their values of SGD $bs$ = 1 (\textit{batch size} = 1), the selectivity of Adadelta $bs$ = 1, 5, 10, and the selectivity of Adam $bs$ = 1 tend to behave similarly to Adagrad where their values begin with a high value and gradually decrease.
Secondly, the starting point of their values changes as the batch size changes. This phenomenon can be clearly observed in SGD and Adam. As the batch size decreases, the starting value of the sparsity and selectivity become higher.
Thirdly, although they tend to arise simultaneously, it is not always the case. This phenomenon has been already introduced in subsection \ref{sparvsselecsection1}.
The sparsity increases while the selectivity decreases in Adadelta $bs$ = 5, 10.

\subsection{Accuracy vs Sparsity and Selectivity 3} \label{accuracyvspsarselec3}
It seems like the sparsity and selectivity decrease across epochs in order to acquire better performance on the test accuracy (See Figure \ref{varyingbatchaccsparselecSM}). This idea is consistent with subsection \ref{accuracyvspsarselec1}: Test accuracy can be improved by reducing sparsity and selectivity. This can be observed in SGD $bs$ = 1, Adadelta $bs$ = 1, 5, 10, and Adam $bs$ = 1. It can be thought of as using more features of digits to increase the accuracy. In other words, using various weights lowers the sparsity and selectivity moving to a better local minimum.
\newline 
\newline 
According to subsection \ref{accuracyvspsarselec2}, it was induced that although the test accuracy, sparsity, and selectivity arise simultaneously, the increasing sparsity and selectivity may hinder the test accuracy. This phenomenon could be observed in Adam $bs$ = 1, 5 (See Figure \ref{varyingbactchsizefigure}). The sparsity and selectivity greatly increased, but the accuracy slightly decreased. The increasing sparsity and selectivity imply that only certain weights get updated. 
This lowers the possibility of finding a new local minimum to increase the test accuracy as it lowers the chance of using various features.
\subsection{Conclusion}
It is evident that the sparsity and selectivity vary because the way to update parameters changes when the batch size varies (See subsection \ref{batch size and GA} to check how varying the batch size could affect the parameter updating rule). Several conclusions could be drawn from this experiment. 
Firstly, the sparsity and selectivity could be significantly influenced by simply varying the batch size. For example, considering that SGD with $bs$ = 1000 showed the worst performance in the test accuracy, sparsity, and selectivity, it was possible to outperform (or perform similar to) other adaptive optimizers simply by changing \textit{bs} = 1.
Secondly, there is an optimal batch size for each optimizer to acquire the highest sparsity and selectivity. 
\subsection{Limitations}
Generalization ability is measured by test accuracy. Although it is not wrong, it would be better if the train loss and test loss could be taken into account, as well as test accuracy, to measure the generalization ability. (See subsection \ref{generalisationabilityloss} for more details about how loss could be used to measure generalization ability.).
This would produce a more robust relationship between generalization performance, sparsity, and selectivity. This not only applies to this experiment; but also other experiments in this paper.
\subsection{Additional Notes}
This paper stated that if the batch size is small, the latest gradient can be used to update parameters, resulting in better generalization performance. It can be thought that high sparsity and selectivity can be aroused by taking the recent past gradient into account when updating parameters. 
This can be considered a similar argument to the conclusion of section \ref{varyinghyperparamlabel} where this paper stated: The more past gradient is taken into account when updating parameters (As the value of $\gamma$ and $\rho$ increases), the higher the value of the sparsity and selectivity gained. 
However, it is an absolutely different meaning. A small batch size vs large batch size in terms of SGD with momentum can be expressed by the following equation:
\newline 
\begin{equation} \label{sgdwithmomentum}
\theta_{t+1,i} = \theta_{t,i} - \alpha \cdot v_{t,i}
\end{equation}
where small batch 
$$v_{t,i} = \gamma \cdot v_{t-1,i} + {\sum_{j=0}^{n-1} {\sum_{i=1}^{m} \nabla_{\theta} L_{i+jm} (\theta_{k+j})}}$$
$$v_{-1,i} = 0$$
where large batch, increasing the batch size by $n$ times:
$$v_{t,i} = \gamma \cdot v_{t-1,i} + {\sum_{i=1}^{nm} \nabla_{\theta} L_{i} (\theta_{k})}$$
$$v_{-1,i} = 0$$
(Other adaptive optimizers could also be reformulated like the above by comparing the small batch and large batch) $\gamma$ is related with $v_{t-1,i}$ term and the batch size is related with $\sum$ term; therefore, the two implies a different meaning.

\section{Extension of Varying the Batch Size}
\label{extensionofvaryingbatchsize}
\subsection{Experimental details}
This experiment tries to answer the following question: Does a smaller batch size really help the performance of the sparsity and selectivity in SGD and Adam? In order to answer the question, the batch size is fixed to 1, and when training the neural network, the data will not be trained in a random order as the baseline. This section implements the following three experiments: 1. \textit{batch size} = 1 with random order, 2. \textit{batch size} = 1 with 5 same consecutive digits, and \textit{3. batch size} = 1 with 10 identical consecutive digits. By experimenting with these, it would be possible to reinforce the conclusion from section \ref{varyingbatchsize}.
\begin{table}[h!]
\centering
 \begin{tabular}{|| c | c | c | c ||} 
 \hline
 Name & Same consecutive numbers & batch size & hyper-parameters \\ [0.5ex] 
 \hline
 \hline
 baseline & - & 50 & $\cdot$\\
 \hline 
 con 1 & 1 & 1 & $\cdot$\\
 \hline 
 con 5 & 5 & 1 & $\cdot$\\
 \hline 
 con 10 & 10 & 1 & $\cdot$\\
 \hline 
\end{tabular}
\caption{The experiment setting for all four optimizers. $\cdot$ indicates the baseline model with the default hyper-parameter setting. $-$ indicates the baseline model with \textit{batch size} = 50. `con 1' indicates it is \textit{batch size} = 1 without any extra manipulation to the data. An example of `con 5' is as follows: 3 $\rightarrow$ 3 $\rightarrow$ 3 $\rightarrow$ 3 $\rightarrow$ 3  $\rightarrow$ 9 $\rightarrow$ 9 $\rightarrow$ 9 $\rightarrow$ 9 $\rightarrow$ 9 $\rightarrow$ 7 $\rightarrow$ 7 $\rightarrow$ 7 $\rightarrow$ 7 $\rightarrow$ 7 $\dots$. An example of `con 10' is as follows: 3 $\rightarrow$ 3 $\rightarrow$ 3 $\rightarrow$ 3 $\rightarrow$ 3 $\rightarrow$ 3 $\rightarrow$ 3 $\rightarrow$ 3 $\rightarrow$ 3 $\rightarrow$ 3 $\rightarrow$ 7 $\rightarrow$ 7 $\rightarrow$ 7 $\rightarrow$ 7 $\rightarrow$ 7 $\rightarrow$ 7 $\rightarrow$ 7 $\rightarrow$ 7 $\rightarrow$ 7 $\rightarrow$ 7 $\rightarrow$ $\dots$.}
\label{extensionvaryingbatchsize}
\end{table}

\subsection{Results}
\begin{figure}[h!]
  \centering
  \includegraphics[width=1.0\textwidth]{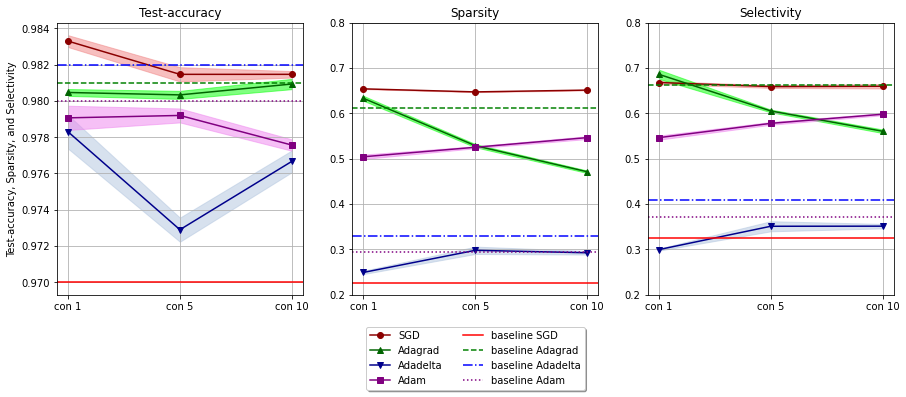}
  \caption{Test accuracy, sparsity, and selectivity. See Figure \ref{batchsize1allSM} for test accuracy, sparsity, and selectivity across epochs. Adagrad and Adadelta will not be discussed as this experiment focuses on SGD and Adam. Data represented as mean and standard error (n=3). }
  \label{batchsize1}
\end{figure}
For all experiments with \textit{batch size} = 1, SGD shows a better or similar sparsity and selectivity. Adam \textit{batch size} = 1 shows lower performance on the test accuracy than the baseline Adam. However, it shows much better performance on the sparsity and selectivity. It can be concluded that a smaller batch size certainly gives rise to the high performance of the sparsity and selectivity in SGD and Adam. 
In addition, it can be derived that inputting the same consecutive digits does not help in the performance of the test accuracy by observing SGD and Adam.
\subsection{Conclusions}
Lowering the batch size assuredly helps the sparsity and selectivity to arise in SGD and Adam. The generalization ability of SGD con 1 outperforms the other optimizers. 
This fact supports the argument of Keskar and Socher and Chen et al. \cite{keskar2017improving, chen2018closing}, where they noted adaptive method might harm generalization performance and Wilson et al. and Zhou et al. \cite{wilson2017marginal, zhou2020towards}, where they stressed SGD sometimes performs better in terms of generalization ability. 
At the same time, SGD con 1 shows a better or similar performance than any other experiments in terms of sparsity and selectivity. 

\section{Varying the number of neurons}
\label{numneuronslabel}
\subsection{Experimental details}
This section examines test accuracy, sparsity, and selectivity of each optimizer by varying the number of neurons in the hidden layer. 
If there is a notable difference in them while varying the number of neurons, I will analyse why such a phenomenon occurs. 
For simplicity, no more than one hidden layer is applied. 
Inspired by the fact that there is a relationship between the batch size, sparsity, and selectivity, this section tries not only \textit{batch size} = 50; but also \textit{batch size} = 1. 
If it turns out that the sparsity and selectivity increase for \textit{batch size} = 1 compared to \textit{batch size} = 50, this paper can support the previous argument: \textit{batch size} = 1 increases the sparsity and selectivity in SGD and Adam. The detailed experimental details can be found through the following table.
\begin{table}[h!]
\centering
 \begin{tabular}{||c | c | c | c||} 
 \hline
 The number of neurons in a single hidden layer & batch size\\ [0.5ex] 
 \hline
 \hline
 [64, 128, 256, 384, 512, 640, 768] & 1\\
 \hline 
 [64, 128, 256, 384, 512, 640, 768] & 50\\
 \hline 
\end{tabular}
\caption{Varying the number of neurons in the single hidden layer. The default hyper-parameters were used. Epoch is set to 30.}
\label{varyingneuronstable}
\end{table}
\subsection{Results and Analysis}
\begin{figure}[h!]
  \centering
  \includegraphics[width=1.0\textwidth]{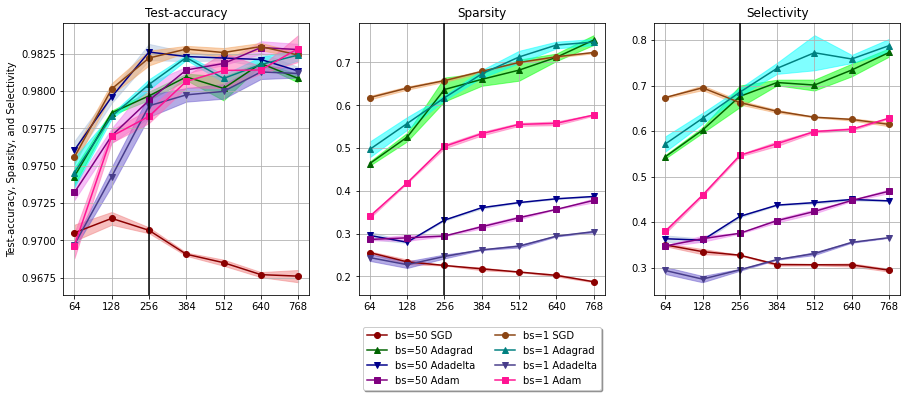}
  \caption{Test accuracy, sparsity, and selectivity by varying the number of neurons. See Figure \ref{number of neurons b=50} and \ref{number of neurons b=1} for more details. This figure shows the extracted test accuracy, sparsity, and selectivity at the last epoch; whereas Figure \ref{number of neurons b=50} and \ref{number of neurons b=1} show their performance across epochs. Data represented as mean and standard error (n=3).}
  \label{numneurons}
\end{figure}
\begin{table}[h!]
\centering
 \begin{tabular}{||c | c | c | c | c | c | c | c||} 
 \hline
 Number of neurons & 64 & 128 & 256 & 384 & 512 & 640 & 768\\ [0.5ex]
 \hline
 \hline
 Number of parameters & 50890 & 101770 & 203530 & 305290 & 407050 & 508810 & 610570\\
 \hline 
\end{tabular}
\caption{The number of neurons and the corresponding number of parameters}
\label{numberofparameters}
\end{table}
In all experiments except SGD \textit{bs} = 50 (\textit{batch size} = 50), the test accuracy tends to increase when the number of neuron increases. 
Also, the sparsity and selectivity increase when the number of neurons increases. 
It may be possible to explain why this phenomenon occurs by the following: if there are many neurons, it provides an opportunity to the neural network to reduce overlap between parameters. 
In other words, neurons are less likely to interfere with each other. 
Also, it would be easier to save notable features of digits in each neuron by increasing the neurons. 
The neurons can have more specific features of each digit.
This would encourage the neural network to be more sparse and selective.
\newline 
\newline 
All experiments tend to perform well in terms of test accuracy when the number of neuron increases in the hidden layer except SGD \textit{bs} = 50\footnote{It is likely that increasing the number of neurons with \textit{batch size} = 50 in SGD over-fits the training data due to over-parameterisation. According to Neyshabur et al. \cite{neyshabur2017geometry}, too many neurons in hidden layers would incur over-fitting, losing generalization ability. This is so called approximation-estimation trade-off behaviour.}. 
This is likely to occur because it has been found that over-parameterisation\footnote{The over-parameterisation is closely related to the number of neurons in the hidden layer. 
As the number of neurons increases, it becomes over-parameterised. See table \ref{numberofparameters}.} in neural network improves generalization performance \cite{allen2018learning, chang2020provable}.
Allen-Zhu et al. \cite{allen2018learning} mathematically proved that why this is likely to have happened in SGD. 
The result of this experiment supports their argument: ``Overparameterisation improves generalization performance". 
\newline
\newline 
It can be seen that the neural network naturally tends to increase the sparsity when the number of neurons increases. 
This happens not only in SGD; but also in other adaptive optimizers. The naturally increasing sparsity indicates that the network naturally reduces over-parameterisation effect. 
In other words, although there are numerous parameters in the network, the network naturally turns off some of their effect. This effect is likely to lead the neural network to be sparse, preventing over-fitting result in better generalization performance \cite{hebiri2020layer}. 
Moreover, the increment of a neuron allows sophisticated computation leading to better generalization ability. At first glance, the explanation about the spareness of the network and the sophisticated computation does not seem to conform. 
How is it possible for the network to perform sophisticated calculation if the network is sparse? 
It is perhaps because naturally aroused sparsity selectively turns off the effect of parameters that are meaningless in a parameter update.
Moreover, recent researches from Liu et al. \cite{liu2021sparse, liu2019improving, liu2020learning} show that sparse neural networks can generalize better than their dense counterparts. 
\newline 
\newline 
When comparing \textit{bs} = 50 and \textit{bs} = 1, all optimizers with \textit{bs} = 1 show better or similar performance (in terms of sparsity and selectivity) than \textit{bs} = 50, except Adadelta. In SGD and Adam, it can be seen that there is a big difference between \textit{bs} = 1 and \textit{bs} = 50. In Adagrad, \textit{bs} = 1 and \textit{bs} = 50 shows similar performance. Similar to experiment \ref{varyingbatchsize}, it can be seen that \textit{batch size} = 1 does not help the performance of Adadelta. These facts reinforce the previous conclusion from section \ref{varyingbatchsize}: There is an optimal batch size for acquiring high sparsity and selectivity.
Another thing to note here is the test accuracy, sparsity, and selectivity when SGD \textit{bs} = 1. It can be seen that the test accuracy of SGD \textit{bs} = 1 is always higher or similar to other optimizers. 
At the same time, it shows a relatively the high sparsity and selectivity. 
Such fact reinforces the previous argument from section \ref{varyingbatchsize}: SGD outperforms or perform similar to other adaptive optimizers with a small batch size in terms of the test accuracy, sparsity, and selectivity.
It is difficult to reason why the selectivity decreases in SGD \textit{bs} = 1 when the number of neurons increases, unlike other optimizers. Is it possibly because SGD does not have an adaptive method to update parameters?
\subsection{Conclusion}
As the number of neurons increases, the test accuracy, sparsity, and selectivity tends to increase. 
I argue that although the neural network can be over-parameterised due to the increment of neurons in the hidden layer, the neural network can naturally increase the sparsity and selectivity to reduce overlap between parameters. 
At the same time, the network can perform sophisticated calculations with an increased number of neurons and parameters. These allow acquiring high generalization performance.

\section{Varying the class diversity in a batch} \label{singlepair5numssection}
\subsection{Experimental details} \label{singlepair5numsexperimentaldetails}
This section introduces a new method to train the neural network. 
It is a convention to maintain a diverse classes in the batch.
This experiment breaks the stereotype and puts only specific numbers inside the batch: 1. Training the network by putting only a single class in the batch (e.g. \textit{batch size} = 50, and only 5s in the batch), 2. Training the network by putting a pair of classes in the batch (e.g. \textit{batch size} = 50, and only 3s and 7s in the batch), and 3. Training the network by putting five classes in batch (e.g. \textit{batch size} = 50, and 2s, 4s, 7s, 8s, and 9s in the batch). By comparing the above three and the baseline model, this experiment tries to determine whether varying class diversity in the batch could affect sparsity and selectivity. The experimental details can be found in more detail in the table below.
\begin{table}[h!]
\centering
 \begin{tabular}{||c | c | c | c||} 
 \hline
 Name & How many different digits in the batch & batch size and hyper-parameters\\ [0.5ex] 
 \hline
 \hline
 single & 1 & $\cdot$ \\
 \hline 
 pair & 2 & $\cdot$ \\ 
 \hline 
 5nums & 5 & $\cdot$ \\
 \hline 
 original & random & $\cdot$ \\
 \hline 
\end{tabular}
\caption{Experiment details. $\cdot$ indicates the baseline setting.}
\label{varyingthenumberofdifferentdigits}
\end{table}
\subsection{Results and Analysis}
\begin{figure}[h!]
  \centering
  \includegraphics[width=1.0\textwidth]{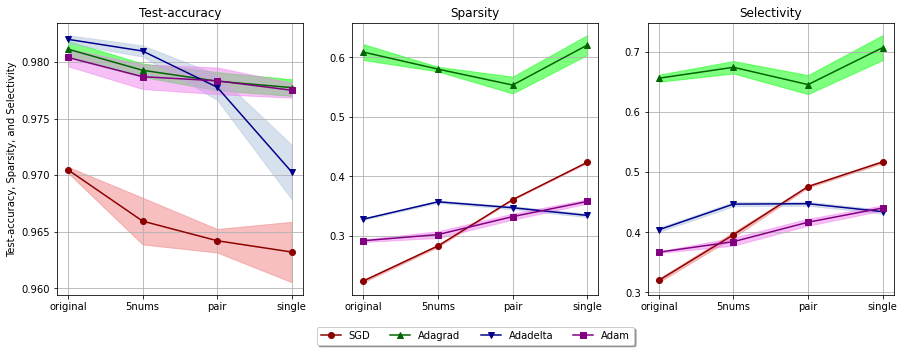}
  \caption{Test accuracy, sparsity, and selectivity of each experiment. See \ref{singlepair5numsSM} for more details. The epoch is set to 30, and values are extracted at the last epoch. Data represented as mean and standard error (n=3).}
  \label{samenumber}
\end{figure}
By observing Figure \ref{samenumber}, reducing the diversity of classes in the batch is considered harmful to the test accuracy. 
As the diversity of digits decreases, the sparsity and selectivity increased in SGD and Adam.
These can be explained by the following: Firstly, lowering diversity in the batch is likely to harm the test accuracy because presenting a similar image to the neural network would stimulate only specific neurons that respond to the particular image feature. 
This likely implies that only certain parameters get updated per iteration, reducing the model's chance to move to a better local optimum. 
Assuming that a dot $A$ on the parameter space $S$ and $A$ is now located in the local minimum in $S$ ($A$ moves by a parameter updating rule), it is hard for $A$ to get away from the minimum when the parameter update is only oriented towards a certain direction. 
It would be easier for $A$ to get away from the local minimum when the parameter update is \textit{not} oriented towards a certain direction (when diverse classes exist in the batch). 
Maintaining diverse classes in the batch would increase the chance for $A$ to move to a better local minimum in $S$. 
Secondly, it was aforementioned that only specific neurons will respond.
As a result, it is likely that only corresponding parameters will get updated per iteration when lowering class diversity in the batch. 
This would increase the sparsity and selectivity. 
\newline
\newline
Combining the first and second explanation, the test accuracy decreases as the sparsity and selectivity increase. This phenomenon also conforms to the conclusion from subsection \ref{accuracyvspsarselec2}: Increased sparsity and selectivity considered harmful to test accuracy. 
There is no clear relationship between the class diversity, the sparsity and selectivity in Adagrad and Adadelta. 
It is hard to reason why the phenomenon of SGD and Adam cannot be applied to Adagrad and Adadelta.
\newline 
\newline 
In the previous experiment, the network was trained by varying the class diversity in the batch. Inspired by the \nameref{extensionofvaryingbatchsize} experiment, another experiment was considered: `\textit{batch size} = 10 + single digit in the batch' (A) vs `\textit{batch size} = 1 + 10 same consecutive digits' (B). By conducting this experiment, it would be possible to observe the influence of reducing the batch size vs reducing class diversity in terms of sparsity and selectivity.
\begin{figure}[h!]
  \centering
  \includegraphics[width=1.0\textwidth]{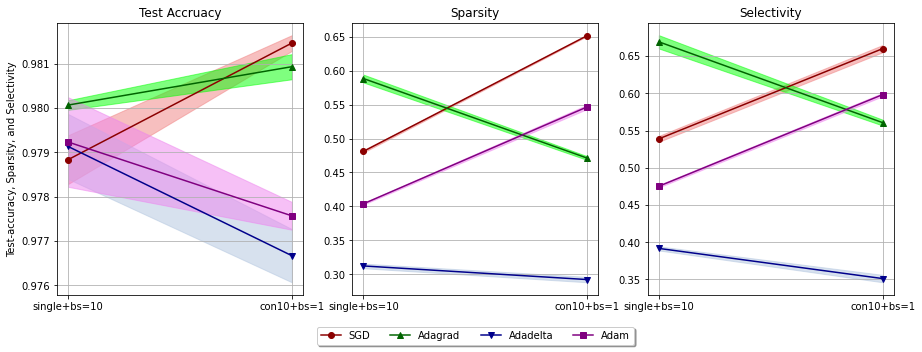}
  \caption{\textit{batch size} = 10 with a single digit in the batch (A) vs \textit{batch size} = 1 with 10 identical consecutive digits (B). Data represented as mean and standard error (n=3)}
  \label{bs10vssub10}
\end{figure}
\newline 
\newline 
The sparsity and selectivity of SGD and Adam increased with condition B. The smaller batch helps the performance of the sparsity and selectivity in SGD and Adam. Setting \textit{batch size} = 1 is more efficient than reducing class diversity in terms of sparsity and selectivity performance in SGD and Adam.
\subsection{Conclusion}
Firstly, lowering class diversity in the batch would help the sparsity and selectivity to arise in SGD and Adam. However, it would impair the generalization performance (See Figure \ref{samenumber}). 
Secondly, although lowering class diversity helps the sparsity and selectivity to arise, setting the \textit{batch size} = 1 provides better performance in SGD and Adam (See Figure \ref{bs10vssub10}). 

\section{Adding a convolutional layer to the neural network}
Convolutional neural network (CNN) is one of the image processing technologies inspired by biological vision \cite{lindsay2020convolutional}. 
It outperforms other machine learning algorithms in classification task and is efficient for feature extraction \cite{jogin2018feature}. 
Feature maps generated by the feature extraction are known to be sparse and selective \cite{cheng2019deep}. Numerous researches used CNN to tackle the classification task.
For example, Coskun et al. and Wang and Li \cite{cocskun2017face, wang2018research} used CNN for face recognition task, and Liu et al. \cite{liu2019real} used CNN for real-time facial expression recognition. 
As CNN is gaining popularity in image processing, this section examines the difference between two models 1. \textit{without CNN} vs 2. \textit{with CNN} in terms of test accuracy, sparsity, and selectivity. 
Yosinski et al. \cite{yosinski2015understanding} found that sparsity was observed in CNN. Rafegas et al. \cite{rafegas2020understanding} mentioned that class selectivity increases as CNN get deepens. However, sparsity and selectivity of FC layer have not been observed by using the metric. This experiment calculates sparsity and selectivity in the FC layer.
\subsection{Experimental details} \label{convexperimentaldetails}
\begin{figure}[h!]
  \centering
  \includegraphics[width=1.0\textwidth]{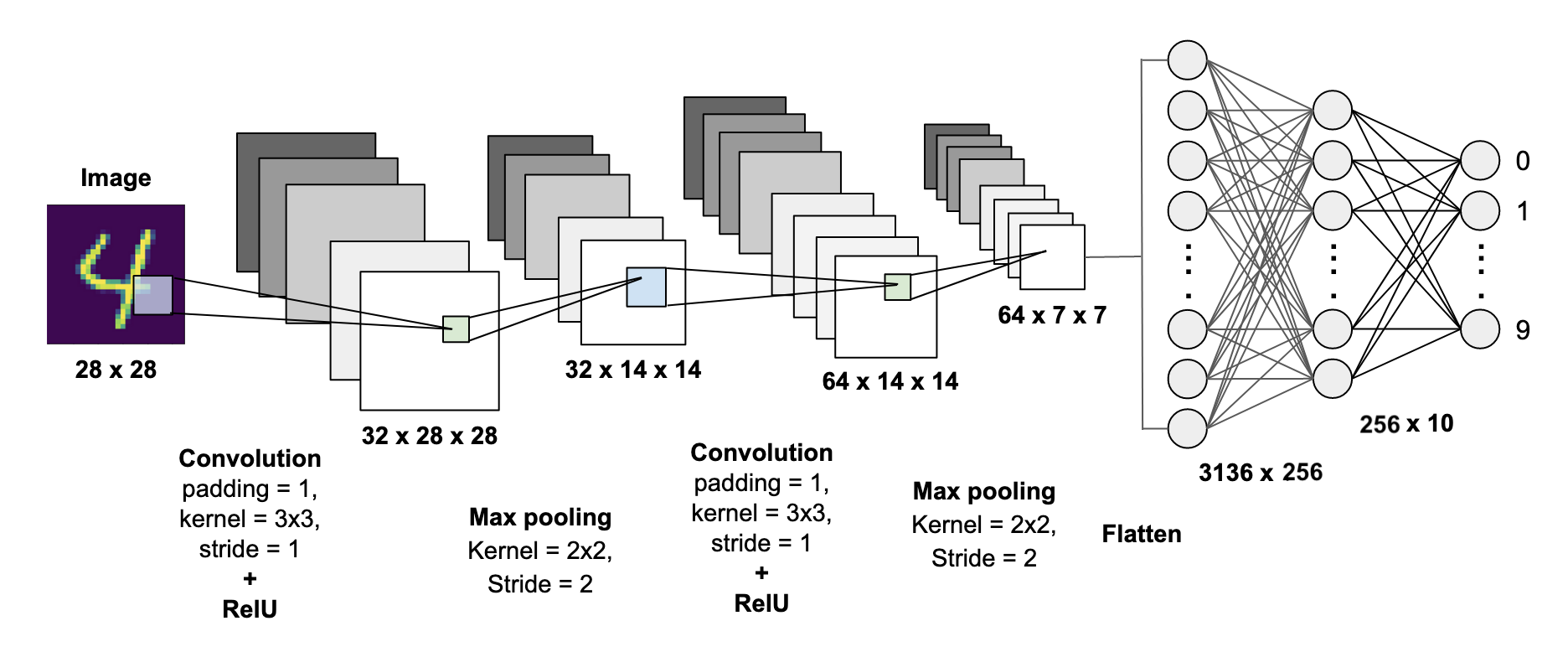}
  \caption{The structure of CNN \cite{patel2019diagram}}
  \label{convlayerstructure}
\end{figure}
\noindent Figure \ref{convlayerstructure} describes the structure of the CNN. In order to maintain consistency with the previous sections, the number of neurons in the FC layer was set to 256. Since the experiment with the CNN reached a certain accuracy promptly, the epoch is set to 5. 
\subsection{Results and Analysis}
\begin{figure}[h!]
  \centering
  \includegraphics[width=1.0\textwidth]{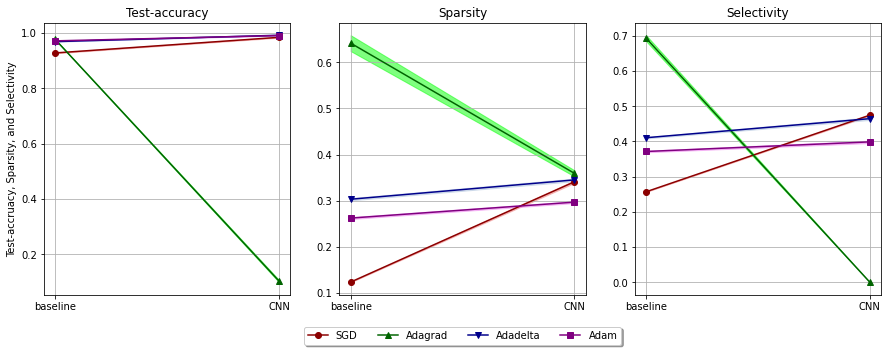}
  \caption{Test accuracy, sparsity, and selectivity: with CNN vs without CNN. See Figure \ref{convallSM} for test accuracy, sparsity, and selectivity across epochs. Data represented as mean and standard error (n=3)}
  \label{convallfig}
\end{figure}
\noindent By observing Figure \ref{convallfig}, it can be seen that the test accuracy, sparsity, and selectivity increased when the CNN is applied to the baseline neural network except for Adagrad\footnote{The test accuracy of Adagrad is very low because this CNN structure is not appropriate to Adagrad. It would be possible to acquire better test accuracy by using another CNN structure.}. It can be interpreted that CNN helped the neural network to extract essential features from the image \cite{hossain2019recognition, rahman2015bangla}, increasing the sparsity and selectivity in the FC layer. 
\subsection{Accuracy vs Sparsity and Selectivity 4}  \label{accuracyvspsarselec4}
In Adagrad, although the sparsity of Adagrad CNN is between 0.3 and 0.4, the selectivity of Adagrad is very poor (It is nearly 0). This evidence supports the argument: Sparsity does not always yield selectivity \cite{spar_selec_does_not_related} and the previous results from subsection \ref{sparvsselecsection1} and \ref{sparvsselecsection2}. 
\newline 
\newline 
Although the test accuracy of Adagrad CNN is very poor, the sparsity of Adagrad CNN maintains a specific value around 0.35.
This might indicate that neural network tends to orient towards a certain degree of sparsity regardless of test accuracy.
\subsection{Conclusion}
Through this experiment, it was found that the test accuracy, sparsity, and selectivity of the FC layer increased when the appropriate convolutional layer was added to the network. The sparsity and selectivity of Adagrad did not increase because the test accuracy was very poor. 

\section{Sorted vs Unsorted} \label{sortedvsunsortedlabel}
\subsection{Experimental details}
In this section, this paper will compare the baseline model and the sorted MNIST data-set model and evaluate which method is better in terms of sparsity and selectivity. 
In the baseline model, the data was shuffled and trained. In other words, when training, the data is trained in random order. The sorted data-set means that the 60,000 training data were sorted in the order of the digits\footnote{Only one number can be placed in the batch. A total of nine batches is mixed with another digit such as 0/1 and 2/3. This will not have a significant impact on the training.}.
\begin{table}[h!]
\centering
 \begin{tabular}{||c | c | c | c||} 
 \hline
 Name & sorted? & epochs & batch size and hyper-parameters\\ [0.5ex] 
 \hline
 \hline
 baseline & No & 100 & $\cdot$\\
 \hline
 sorted & Yes & 100 & $\cdot$\\
 \hline
\end{tabular}
\caption{Experimental details of baseline model vs sorted. $\cdot$ indicates that the \textit{batch size} = 50 and it uses the baseline hyper-parameters.}
\label{sortvsunsortbaseline}
\end{table}
\subsection{Result}
\begin{figure}[h!]
  \centering
  \includegraphics[width=1.0\textwidth]{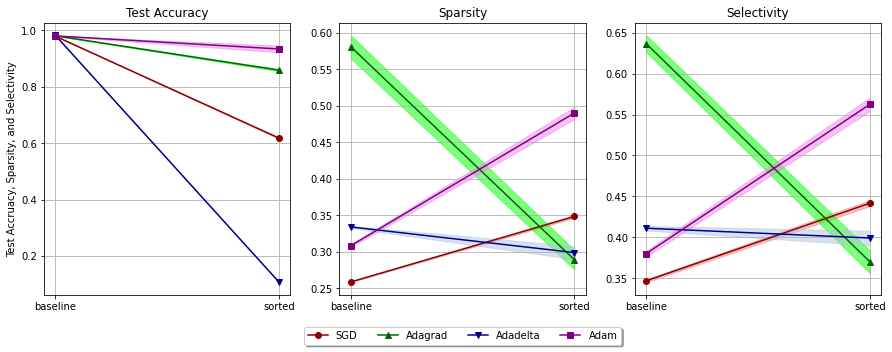}
  \caption{Test accuracy, sparsity, and selectivity of unsorted(baseline) vs sorted model. See Figure \ref{sortedvsunsortedallSM} for more details. The epoch is set to 100. Data represented as mean and standard error (n=3).}
  \label{sortedvsunsorted}
\end{figure}
By observing Figure \ref{sortedvsunsorted}, the test accuracy decreased in \textit{sorted}. This is likely to happen because training with \textit{unsorted} (baseline) data-set is more likely to bounce out from the local minimum better than \textit{sorted}. 
Training \textit{sorted} is unlikely to easily get away from the local minimum since the parameter update will be constantly oriented towards a very similar direction because only certain neurons respond (certain parameters respond) for a certain extent of iterations. This explanation could also explain why \textit{sorted} shows better performance on the sparsity and selectivity than \textit{unsorted} in SGD and Adam\footnote{This explanation is very similar to section \ref{singlepair5numssection}. A similar tendency could be observed in the test accuracy, sparsity, and selectivity in SGD and Adam (See Figure \ref{samenumber}).}.
In Adagrad and Adadelta \textit{baseline} shows better performance on the sparsity and selectivity than \textit{sorted}. 
While the test accuracy decreases among all optimizers, the sparsity and selectivity does not have a constant tendency. It is difficult to infer why this phenomenon occur.
\subsection{Accuracy vs Sparsity and Selectivity 5}  \label{accuracyvspsarselec5}
An unusual phenomenon could be observed in sorted Adadelta. Although the test accuracy is very low, the sparsity and selectivity is not very low. Rather, they maintain a certain degree of values similar to other optimizers. This result is similar to subsection \ref{accuracyvspsarselec4}. This may indicate that the neural network tends to orient towards a certain degree of sparsity and selectivity regardless of test accuracy.
\subsection{Conclusion}
The sorted data-set is likely to induce a higher sparsity and selectivity compared to the baseline model in SGD and Adam. However, the reduction of the test accuracy should be taken into account. Sorting the data-set does not help the sparsity and selectivity to arise in Adagrad and Adadelta.

\section{A combination of conditions} \label{combinationofconditionslabel}
\subsection{Experimental details}
In the previous experiments, test accuracy, sparsity, and selectivity were examined for each condition. This section discusses them when certain conditions are mixed. It was impossible to investigate all possibilities of mixing conditions because there were too many combinations to consider. 
Assuming there are $k$ conditions, the total possible combinations are $\sum_{i=2}^{k} {\comb{k}{i}}$. In addition, even if all combinations are considered, and the tendency of sparsity and selectivity can be concluded in this paper, there are numerous other conditions that are not taken into account; thus, it is impossible to generalize the tendency. I used two optimizers: SGD and Adam, to discuss the combination of the conditions. It tries to answer the following question:
\begin{enumerate}
\item \textbf{SGD: } Through previous experiments, it has been confirmed that setting \textit{batch size} = 1 and a large momentum hyper-parameter increase the sparsity and selectivity of SGD; therefore, if the two are combined, what would happen to them in the combined model?
\item \textbf{Adam: } Through previous experiments, it has been confirmed that putting a single digit in the batch (reducing class diversity in the batch) and applying a large number of neurons in the hidden layer increase the sparsity and selectivity of Adam; therefore, if the two models are combined, what would happen to them in the combined model?
\end{enumerate}
Detailed experiment details can be checked by the following table.
\begin{table}[h!]
\centering
 \begin{tabular}{||c | c ||} 
 \hline
 Name & Conditions\\ [0.5ex] 
 \hline
 \hline
 SGD combined & SGD batch size = 1 and SGD momentum = 0.9\\
 \hline 
 Adam combined & Adam single and Adam 768 neurons in the single hidden layer\\
 \hline 
\end{tabular}
\caption{Experimental details for the combined model of SGD and Adam}
\label{combination}
\end{table}
\subsection{Results and Analysis}
\begin{figure}[h!]
  \centering
  \includegraphics[width=1.0\textwidth]{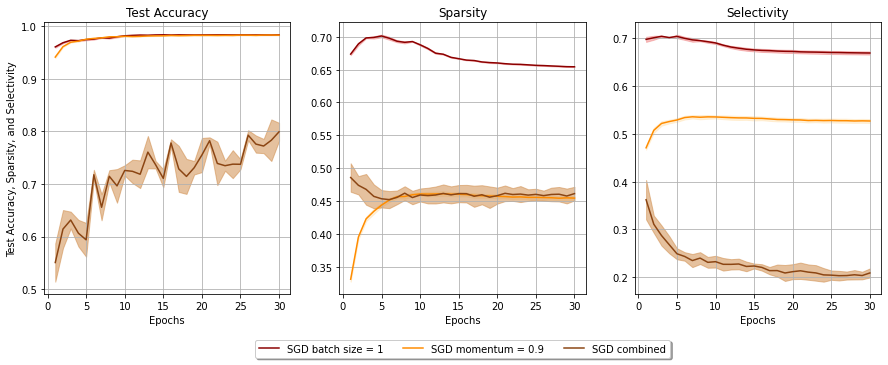}
  \includegraphics[width=1.0\textwidth]{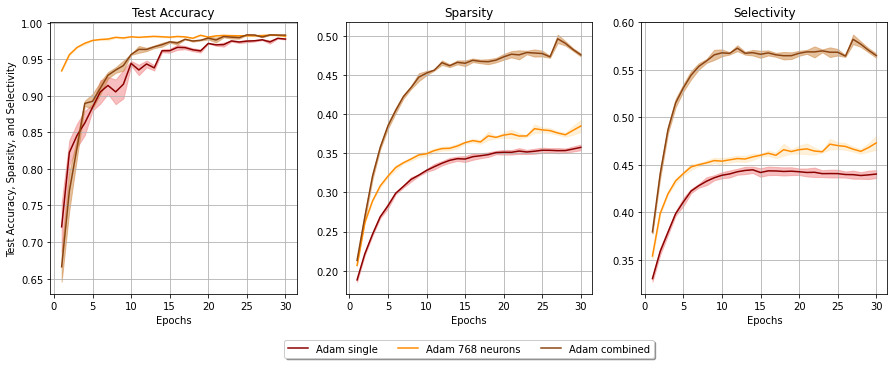}
  \caption{Test accuracy, sparsity, and selectivity of combined models. Top: A combined model of SGD, Bottom: A combined model of Adam. Data represented as mean and standard error (n=3).}
  \label{combinedmodelsSM}
\end{figure}
\begin{enumerate}
\item \textbf{SGD: } When conditions with the high sparsity and selectivity are combined, the test accuracy, sparsity, and selectivity decreased compared to the two models. 
In the previous section, it has been concluded that a smaller batch size produces the high sparsity and selectivity. 
However, it can be seen that it is not always the case. 
The combined model does not perform better than the other two models in the test accuracy and selectivity. It also performs worse than the baseline model in the selectivity. 
Considering this experimental result, two conclusions can be drawn. 
Firstly, the test accuracy, sparsity, and selectivity decrease or remain unchanged when the two high sparsity and selectivity conditions are combined. 
Secondly, a smaller batch size does not always produce better test accuracy, sparsity, and selectivity when a momentum hyper-parameter exists.
\item \textbf{Adam: } By combining the two high sparsity and selectivity conditions, the combined model shows better sparsity and selectivity than the other two models. This contradicts the previous conclusion. The sparsity and selectivity largely improved.
\end{enumerate}
\subsection{Conclusion}
It was found that a smaller batch size does not always yield better test accuracy, sparsity, and selectivity when a high value of momentum hyper-parameter is applied to SGD. 
This seems to be the case because the momentum directly affects the parameter update (see equation \ref{sgdwithmomentum}). 
Also, by analysing the difference between the combined SGD model and combined Adam model (SGD shows a worse performance when combining the high sparsity and selectivity conditions, and Adam offers better performance when integrating the high sparsity and selectivity conditions), it can be seen that when one particular condition meets the other condition, they can have a synergy effect on the performance (Adam), or they can harm the performance (SGD); hence, it would be possible to combine conditions to yield better sparsity and selectivity.

\chapter{Conclusion}
\label{chap:conclusion}
This paper examined many conditions that influence sparsity and selectivity of neural network. 
There were mainly eight approaches: Varying the number of hidden layers, varying the hyper-parameters, varying the batch size, varying the number of neurons in a hidden layer, varying the class diversity in a batch, applying a convolutional layer, comparing sorted vs unsorted data-set, and a combination of conditions. 
This research aims to discover \textit{which conditions naturally increase or decrease sparsity and selectivity}, not \textit{why sparsity and selectivity increase or decrease}. 
Nevertheless, I included why sparsity and selectivity vary under a particular condition by relating them to generalization ability or parameter updating methods. 
This chapter will re-summarise all the findings and present limitations by using bullet points. In addition, future works will be discussed. 
\section{Summary of all findings}
\begin{figure}[h!]
  \centering
  \includegraphics[width=1.0\textwidth]{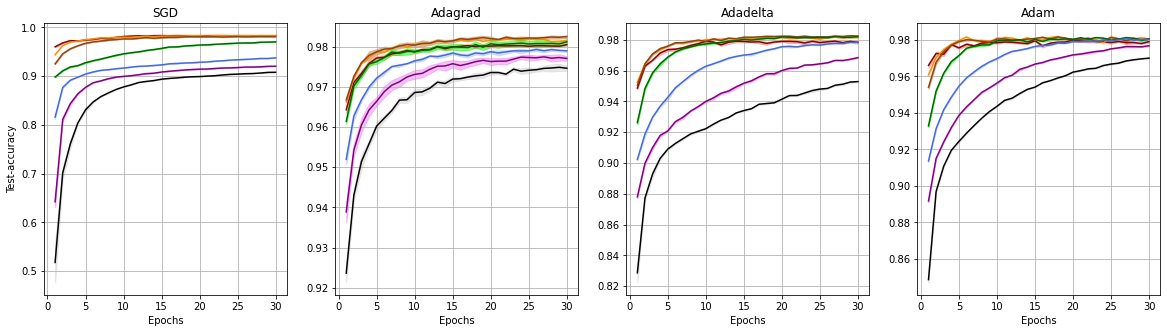}
  \includegraphics[width=1.0\textwidth]{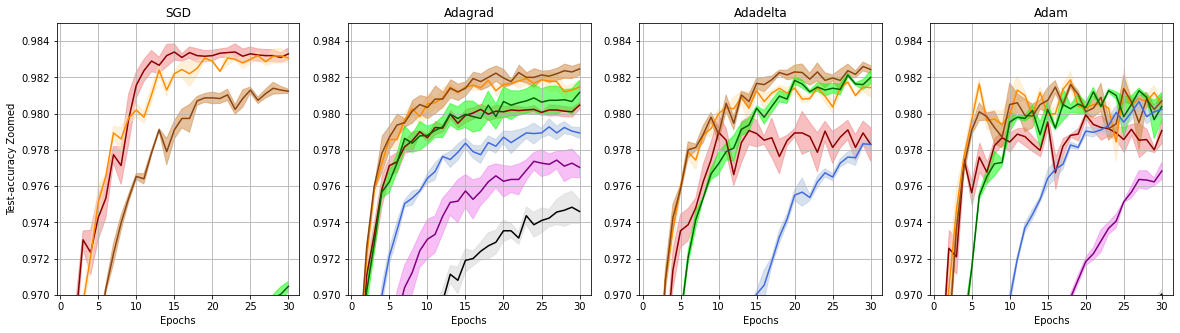}
  \includegraphics[width=1.0\textwidth]{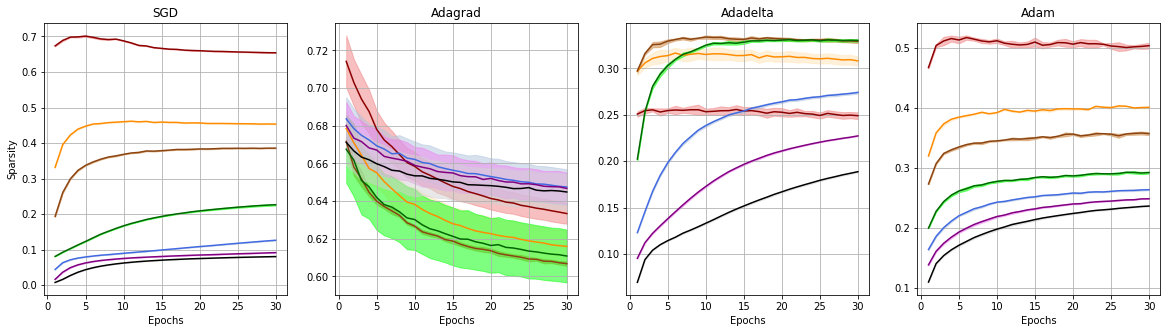}
  \includegraphics[width=1.0\textwidth]{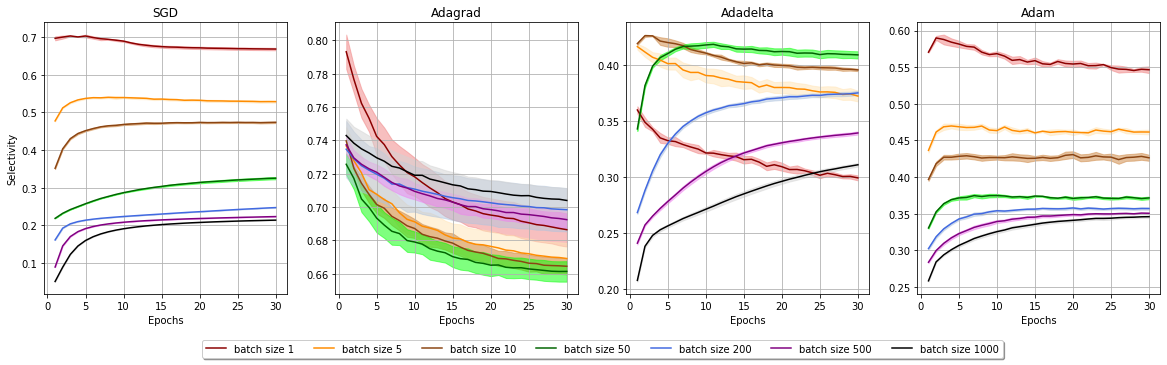}
  \caption{Test accuracy, sparsity, and selectivity by varying the batch sizes. Data represented as mean and standard error (n=3)}
  \label{varyingbatchaccsparselecSM}
\end{figure}

\begin{figure}[h!]
  \centering
  \includegraphics[width=1.0\textwidth]{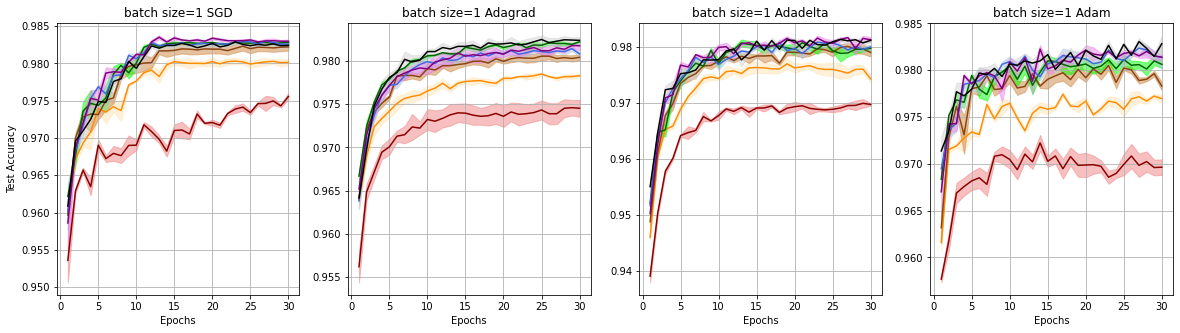}
  \includegraphics[width=1.0\textwidth]{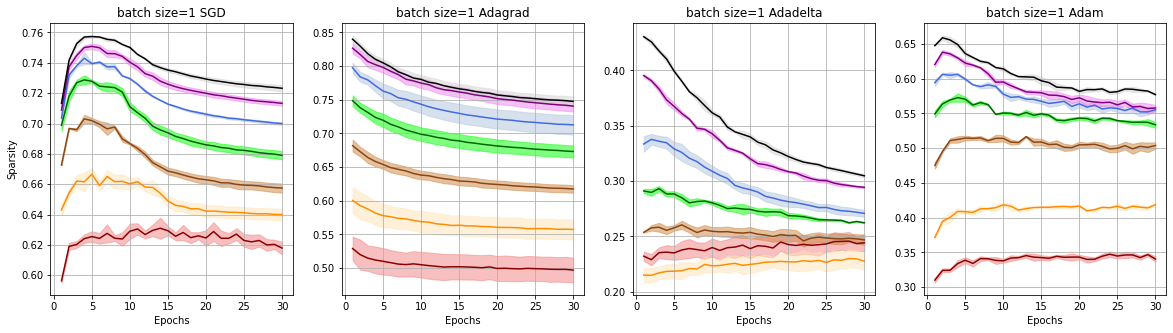}
  \includegraphics[width=1.0\textwidth]{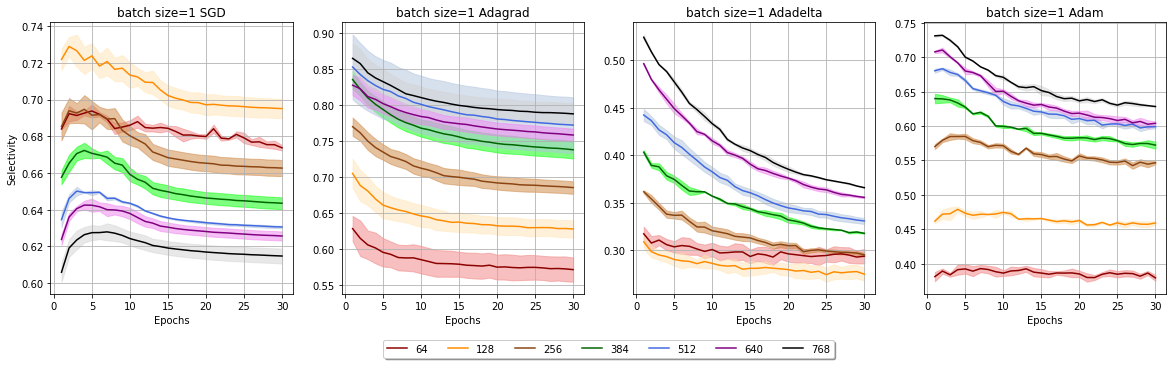}
  \caption{Test accuracy, sparsity, and selectivity when varying the number of neurons in the single hidden layer with \textit{batch size} = 1. Data represented as mean and standard error (n=3)}
  \label{number of neurons b=1}
\end{figure}

\begin{figure}[h!]
  \centering
  \includegraphics[width=1.0\textwidth]{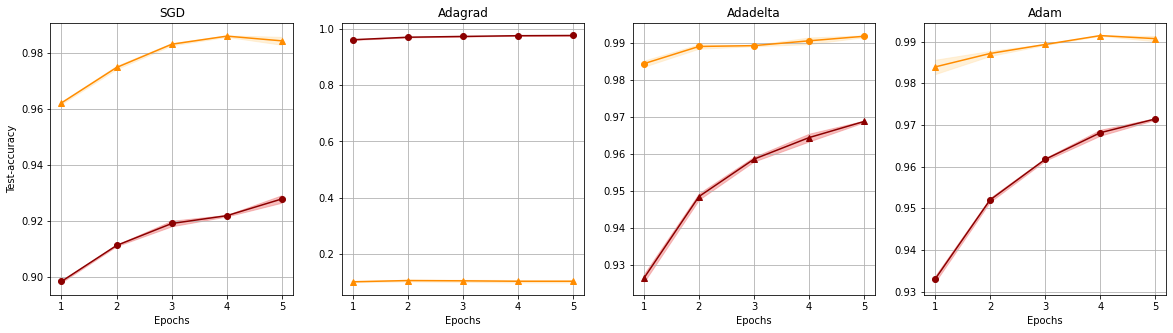}
  \includegraphics[width=1.0\textwidth]{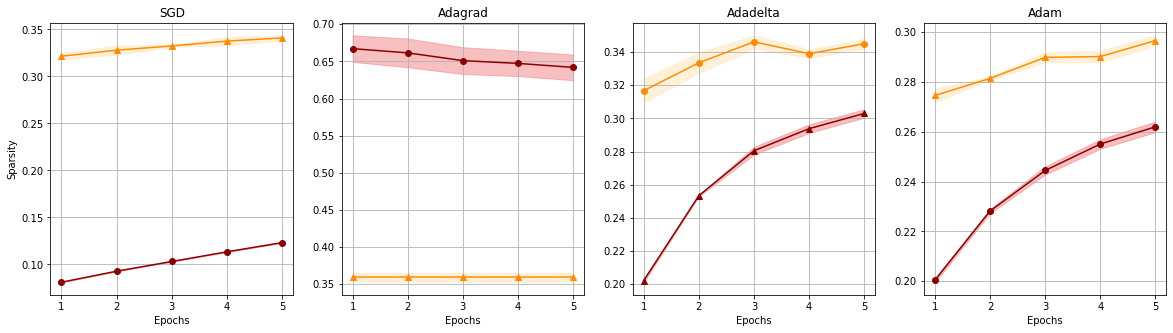}
  \includegraphics[width=1.0\textwidth]{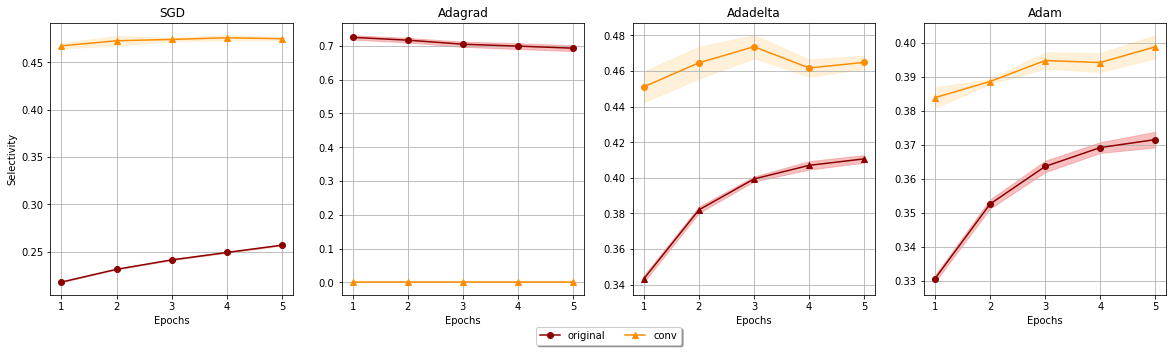}
  \caption{Test accuracy, sparsity, and selectivity comparing the baseline model and baseline model with conv layer. See \ref{convexperimentaldetails} for the convolutional layer structure. Data represented as mean and standard error (n=3)}
  \label{convallSM}
\end{figure}

\begin{figure}[h!]
  \centering
  \includegraphics[width=1.0\textwidth]{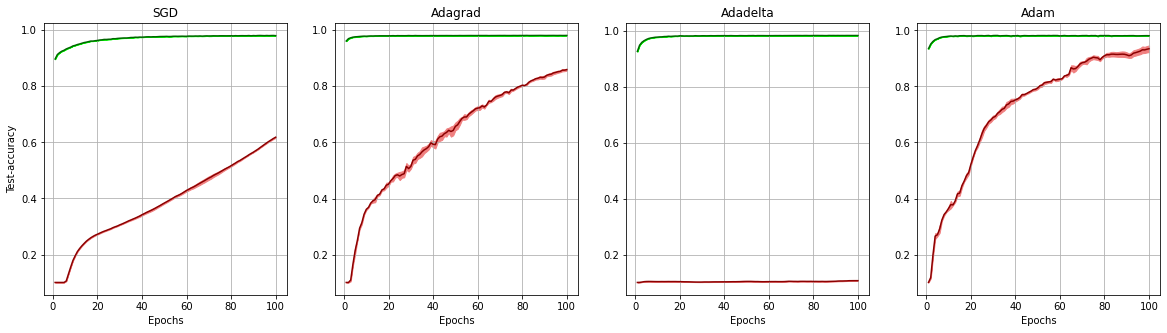}
  \includegraphics[width=1.0\textwidth]{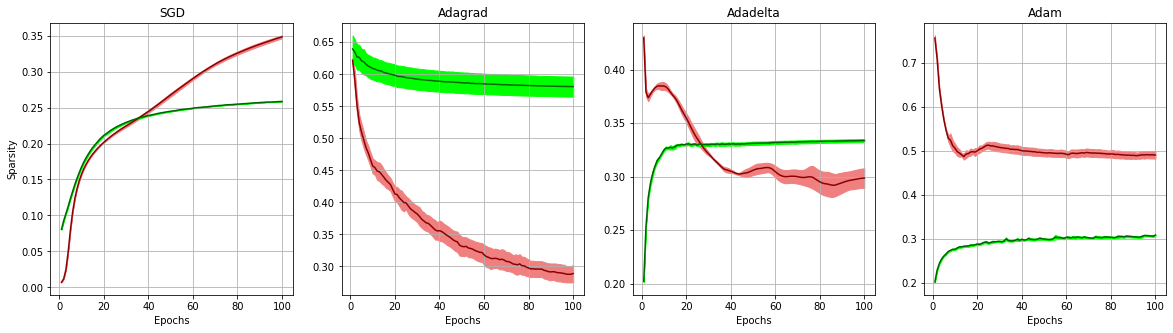}
  \includegraphics[width=1.0\textwidth]{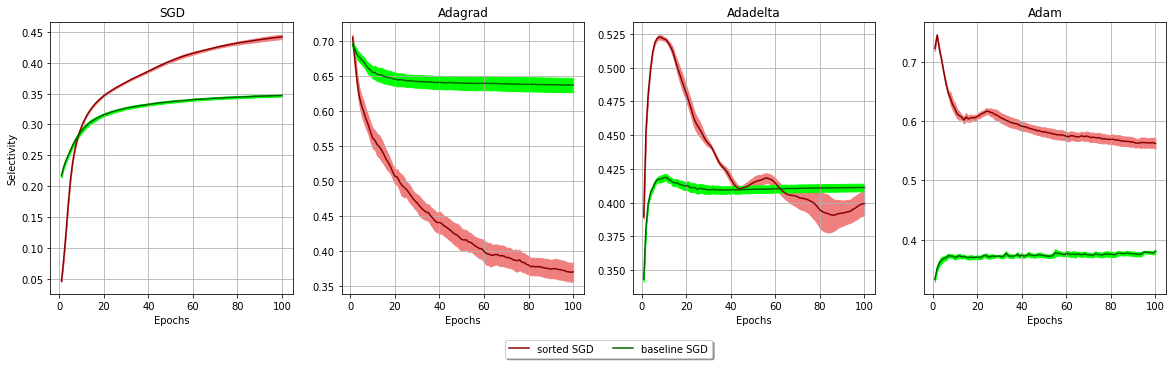}
  \caption{Test accuracy, sparsity, and selectivity comparing sorted vs unsorted. Data represented as mean and standard error (n=3)}
  \label{sortedvsunsortedallSM}
\end{figure}

\begin{itemize}

    \item The tendency of the sparsity and selectivity across epochs depends on optimizers (See Figure \ref{baselinemodelsparandselec}).
    
    \item The tendency of varying the sparsity and selectivity across epochs is altered by changing the condition. For example, the sparsity and selectivity of SGD, Adadelta, and Adam baseline models begin with a low value and gradually increase. Their values of the Adagrad baseline model begin with a high value and gradually decreases. These tendencies could be changed by varying the batch size (See Figure \ref{varyingbatchaccsparselecSM}).
    
    \item The condition yields different responses of the sparsity and selectivity depending on the optimizers. For example, condition A makes sparsity increase in optimizer X. Conversely, condition A makes sparsity decrease in optimizer Y. This is likely to happen because each optimizer uses different parameter update methods (See Figure \ref{varyingbactchsizefigure}).
    
    \item The sparsity and selectivity not only gradually increase or decrease across epochs; rather, they increase, decrease, or fluctuate until they reach a stable value (See SGD in Figure \ref{number of neurons b=1}).
    
    \item There is a slight difference between training and testing in terms of sparsity and selectivity values. The difference between them decreases as the epoch progresses (See Figure \ref{trainingtesting} and \ref{trainingtesting2}).
    
    \item The sparsity and selectivity normally tend to arise simultaneously. However, there are exceptions. It has been observed that although the sparsity increases, the selectivity decreases in both across epochs (e.g. Adadelta in Figure \ref{number of neurons b=1}) and across the last epoch (e.g. SGD and Adam in Figure \ref{hyperhyper}). 
    
    \item The test accuracy, sparsity, and selectivity usually tend to arise simultaneously. However, there are exceptional cases. After reaching a certain degree of the test accuracy, a neural network decreases the sparsity and selectivity to obtain better generalization ability (See Figure \ref{varyingbatchaccsparselecSM} and \ref{number of neurons b=1}). Also, it has been observed that increasing the sparsity and selectivity in the network is likely to harm generalization ability (See Figure \ref{samenumber}). The exceptional cases arise randomly.
    
    \item The test accuracy is sometimes unaffected by varying the sparsity and selectivity (See Figure \ref{accsparselec}).
    
    \item The sparsity and selectivity cannot predict the test accuracy, and vice versa.
    
    \item It has been observed that a neural network maintains a certain degree of the sparsity (and selectivity) regardless of the test accuracy (See Adadelta in Figure \ref{sortedvsunsortedallSM} and Adagrad in Figure \ref{convallSM}). According to Arora et al. \cite{arora2018stronger}, over-parameterised networks could be compressed to simpler networks with a reduced number of parameters without compromising their ability to generalize. 
    Nonetheless, training simpler networks directly can lead to negative result \cite{arora2018stronger}. 
    This may be explained by the fact that neural network aims at a certain degree of sparsity in any situation. 
    Even if it is compressed, it is possible to sustain generalization ability because the network itself is already sparse. 
    In other words, essential parameters that affect learning does not get influenced by compressing the network. 
    However, it would not be possible to train a simpler network (S) from the beginning because S also wants a certain degree of sparsity regardless of any situation. 
    For example, let's assume 60 parameters are needed to compute task T. 
    If there are 100 parameters and S wants 50\% of the sparsity, S cannot perform the task T because S needs ten more parameters.
    
    \item Increasing the number of hidden layers is likely to help the sparsity and selectivity to arise in Adadelta and Adam (See Figure \ref{ssvsnumberofhiddenlayer}).
    
    \item The sparsity and selectivity are likely to increase or fluctuate by passing through the hidden layer. This phenomenon could be observed in SGD and Adadelta. There is no clear tendency in Adam. They rather decrease by going through the hidden layer in Adagrad (See Figure \ref{hl_each_extracted}).
    
    \item Adding the weight decay hyper-parameter is considered harmful to the sparsity and selectivity in all optimizers (See Figure \ref{wdhyper}).
    
    \item The optimal learning rate for the highest test accuracy does not guarantee the highest sparsity and selectivity; instead, they arise when the learning rate is slightly higher than the optimal learning rate. Sometimes, the optimal learning rate for the highest sparsity and selectivity is different. In other words, the learning rate $\eta$ yields the highest sparsity when $\eta = \eta_1$ and yields the highest selectivity when $\eta = \eta_2$, $\eta_1 \neq \eta_2$ (See Figure \ref{wdlr}).
    
    \item The more past gradient is taken into account when updating parameters (weights), the higher the sparsity and selectivity value is obtained. The degree of taking the past gradient could be manipulated by $\gamma$ in SGD with momentum, $\rho$ in Adadelta, and $\beta_1$ and $\beta_2$ in Adam. However, considering the past gradient, too much would harm the performance (See Figure \ref{hyperhyper}).
    
    \item There is an optimal batch size for yielding the highest sparsity and selectivity depending on optimizers. The lower the batch size, the higher the sparsity and selectivity are obtained in SGD and Adam (The optimal batch size is 1). However, if a momentum hyper-parameter is added to SGD, the above statement cannot be guaranteed (See Figure \ref{combinedmodelsSM}). The highest sparsity and selectivity are obtained when the batch size is 200 and 50 in Adagrad and Adadelta respectively. 
    
    \item Increasing the number of neurons in a single hidden layer is likely to help the sparsity and selectivity to arise.
    However, there are two exceptional cases. Firstly, increasing the number of neurons with SGD \textit{batch size} = 50 showed a different tendency. The sparsity and selectivity decrease as the number of neurons increases. Secondly, the sparsity increases, but the selectivity decreases with SGD \textit{batch size} = 1 (See Figure \ref{numneurons}).
    
    \item Reducing class diversity in the batch is likely to increase the sparsity and selectivity in SGD and Adam. The reduction in the test accuracy should be considered. Adagrad and Adadelta do not show such tendency (See Figure \ref{singlepair5numssection}). 
    
    \item Adding the convolutional layer to the neural network improves the sparsity and selectivity of the FC layer when a certain degree of the test accuracy could be maintained. Namely, the sparsity and selectivity cannot be increased when adding a convolutional layer harms the test accuracy.
    
    \item Training with the sorted data increases the sparsity and selectivity in SGD and Adam. The reduction in the test accuracy should be considered. Adagrad and Adadelta do not show this tendency (See Figure \ref{sortedvsunsorted}).
    
    \item Two conditions that yield the high sparsity and selectivity could be merged to obtain an even higher sparsity and selectivity. However, they would also produce a worse performance sometimes (See Figure \ref{combinedmodelsSM}).
    
    \item Early stopping\footnote{Early stopping means stopping training before reaching the end of the epoch. Conventionally, early stopping was used to avoid overfitting \cite{bishop2006pattern}. However, if the aim is to acquire the highest sparsity and selectivity, early-stopping could be used in different meaning.} can be used to obtain high sparsity and selectivity (See SGD in Figure \ref{varyingbatchaccsparselecSM}).
    
    \item 
    There may be a close relation between fluctuation, sparsity, and selectivity. The more it fluctuates when training, the higher the sparsity and selectivity. See Figure \ref{wdlr}: SGD with a high learning rate, Figure \ref{hyperhyper}: SGD with a large momentum hyper-parameter, and Figure \ref{varyingbactchsizefigure}: SGD with a small batch size. To my speculation, they can be an indicator of the shape of the local basin as high fluctuation enables the model to converge to a flat minimum. It should be noted that not all flat basins would be better for generalization ability than sharp basins.

\end{itemize}

\section{Limitations}
\begin{itemize}

    \item Gale et al. \cite{gale2019selectivity} argued that CCMAS selectivity might provide misleading estimates. Another selectivity metric can be used for future analysis.

    \item I could not mathematically prove the relationship between the sparsity and selectivity.
    
    \item I could not mathematically prove the relationship between the test accuracy (generalization ability), sparsity, and selectivity.
    
    \item Although it was possible to see the trend of the sparsity and selectivity when varying the number of hidden layers, it would be more robust if more hidden layers could be applied to the neural network (See subsection \ref{hiddenlayerlimitations}).
    
    \item The calculation of the sparsity and selectivity in the \nameref{hiddenlayerlabel} section may not be appropriate.
    
    \item It was deduced in the \nameref{varyinghyperparamlabel} section that the more past gradient is taken into account when updating parameter, the higher value of the sparsity and selectivity is gained. It would be possible to reinforce this argument by experimenting with more optimizers that can control the degree of the past gradient. 
    
    \item It would be possible to derive a more robust relationship between the batch size, sparsity, and selectivity by trying more batch sizes.
    
    \item In the \nameref{numneuronslabel} section, the minimum number of neurons was 64 and the maximum number of neurons was 768. It would be possible to yield a more robust relationship between the number of neurons, sparsity, and selectivity if the minimum number of neurons is set to 32 or less and the maximum number of neurons is set to 1024 or more.
    
    \item In the \nameref{singlepair5numssection} section, there were three experiments: single, pair, and 5nums. It would be possible to acquire a more robust relationship between class diversity in batch, sparsity, and selectivity if there are more experiments, such as 3nums, 4nums, etc.

\end{itemize}

\section{Further works}

\subsection{A Modified version of Adagrad and Adadelta}
When updating parameters (weights) with Adagrad and Adadelta, it is possible to include the square of the gradient. Including the square of the gradient is motivated in biology. The new variant of Adagrad and Adadelta could be expressed by the following:
\begin{enumerate}
\item A new version of Adagrad:
\begin{equation}
    \theta_{t+1,i} = \theta_{t,i} - \frac{\eta \cdot k \cdot {G_{t,i}}^2}{\sqrt{G_{t,i}}+\epsilon} \cdot \nabla_\theta J(\theta_{t,i})
\end{equation}
where:
$$G_{t,i} = G_{t-1,i} + \Big(\nabla_\theta J(\theta_{t,i})\Big)^2$$
\centerline{\textit{k is a new hyper-parameter which controls the effect of the square of the gradient}}
\item A new version of Adadelta:
\begin{equation}
\theta_{t+1,i} = \theta_{t,i} - \eta \cdot k \cdot {(E[g^2]_{t,i})}^2 \cdot \frac{\sqrt{E[\Delta \theta^2]_{t-1,i} + \epsilon}}{\sqrt{E[g^2]_{t,i}+\epsilon}} \cdot \nabla_\theta J(\theta_{t,i})
\end{equation}
where 
$$E[g^2]_{t,i} = \rho E[g^2]_{t-1,i} + (1-\rho)g^2_{t,i}$$
$$E[\Delta \theta^2]_{t-1,i} = \rho E[\Delta \theta^2]_{t-2,i} + (1-\rho)\Delta \theta^2_{t-1,i}$$
\centerline{\textit{k is a new hyper-parameter which controls the effect of the square of the gradient}}
\end{enumerate}
A detailed basic form of these two optimizers could be found in section \ref{optimizerdeepexplain}. Not only these new optimizers are motivated in biology; but also they are related to `no more pesky learning' \cite{schaul2013no}. It would be interesting to analyse test accuracy, sparsity, and selectivity by using these new optimizers and `no more pesky learning' optimizer. The code for the variant of Adagrad and Adadelta could be found on this link: \url{https://github.com/7201krap/Thesis_and_Experiments/blob/main/new_optimizers.ipynb}. These are ready to be used with the new hyper-parameter $k$ tuning.

\subsection{Measuring catastrophic forgetting}
An additional experiment could be executed to see if a highly sparse and selective neural network is certainly beneficial for continual learning. An example experiment could be conducted by the following:
\begin{enumerate}
    \item Generate two models: baseline model and \textit{batch size} = 1 model using SGD optimizer.
    \item Randomly choose 5 digits out of 10 digits. 
    \item Train each model with chosen 5 digits in (2).
    \item Train each trained model acquired in (3) with the rest 5 digits.
    \item Check which model performs better in terms of the test accuracy of firstly chosen 5 digits in (2). 
\end{enumerate}

\subsection{Measuring generalization ability}
\label{generalisationabilityloss}
\begin{figure}[h!]
  \centering
  \includegraphics[width=1.0\textwidth]{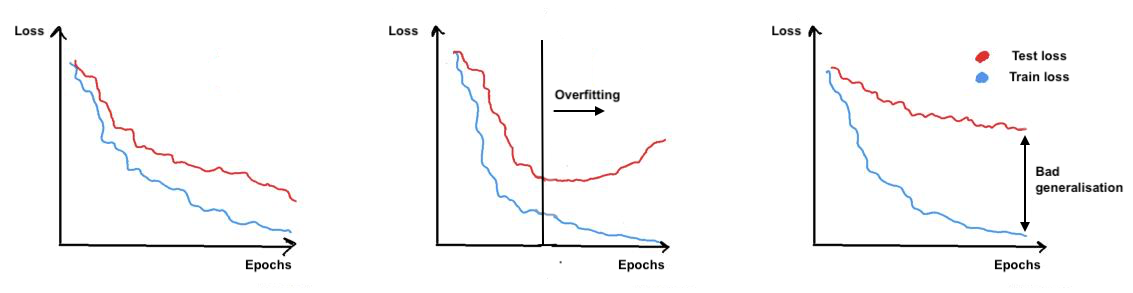}
  \caption{
  Examples of train loss and test loss. The left-hand side plot indicates that training is showing good generalization performance. Both train and test loss decrease as the epoch progresses. The middle plot indicates that the model is over-fitting to train data. It is possible to induce that over-fitting could be avoided by early-stopping. The right-hand side plot demonstrates that they are not showing good generalization performance. The test loss does not get decreased a lot even though the epoch progresses.}
  \label{lossfig}
\end{figure}
\noindent Generalization ability is measured by test accuracy in this paper. This paper explained the behaviour of sparsity and selectivity with regards to test accuracy.
Loss of training and testing could be also utilised to examine generalization performance. 
Using loss in addition to test accuracy might provide better insight into the relationship between generalization ability, sparsity, and selectivity. 

\subsection{A different way to plot graphs}
\begin{figure}[h!]
  \centering
  \includegraphics[width=0.7\textwidth]{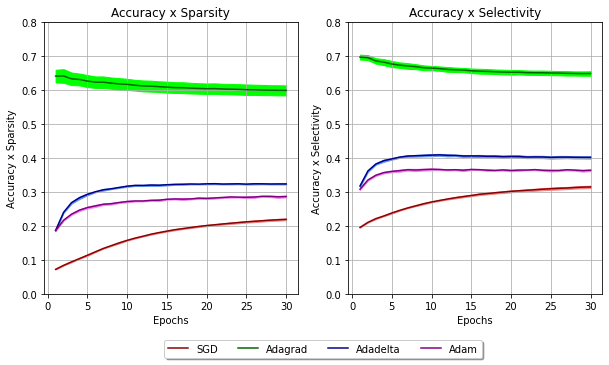}
  \caption{Test accuracy $\times$ sparsity and test accuracy $\times$ selectivity. Data represented as mean and standard error (n=3).}
  \label{sparselec_mul_accuracy}
\end{figure}

Instead of showing three plots: test accuracy, sparsity, and selectivity, it is possible to combine them (e.g. test accuracy $\times$ sparsity and test accuracy $\times$ selectivity) to consider sparsity and selectivity in terms of test accuracy at once. 
However, it turned out that these graphs do not provide enough information. The test accuracy is always very high (nearly 90\% or more); thus, there is seldom a difference between the sparsity and selectivity plot and the test accuracy $\times$ sparsity and test accuracy $\times$ selectivity plot. Compare Figure \ref{sparselec_mul_accuracy} and \ref{baselinemodelsparandselec} to check the difference.
It may be possible to expand this idea and implement this technique for future work.
    
\subsection{Further possible approaches}
\begin{enumerate}

    \item It would be possible to get a more robust relationship between the condition, sparsity, and selectivity by using more optimizers.
    
    \item Other metrics could be used to measure sparsity. See Hurley and Rickard's paper \cite{hurley2009comparing} to check possible sparsity measurements. 
    
    \item This paper used class selectivity to measure selectivity. Similarly, colour selectivity could be measured if the data is not a gray-scale image. See Rafegas et al.'s paper \cite{rafegas2020understanding} to check colour selectivity measurement.
    
\end{enumerate}





\bibliography{dissertation}

\chapter{Supplementary Materials (SM)}  \label{SM}

\begin{figure}[h!]
  \centering
  \includegraphics[width=1.0\textwidth]{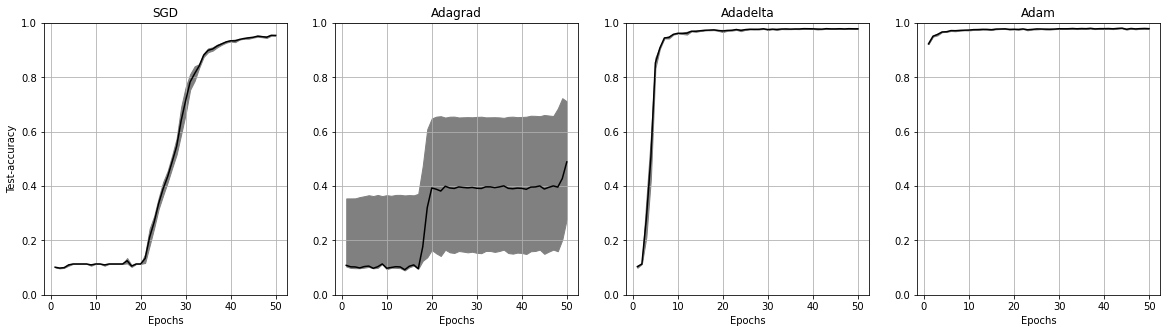}
  \includegraphics[width=1.0\textwidth]{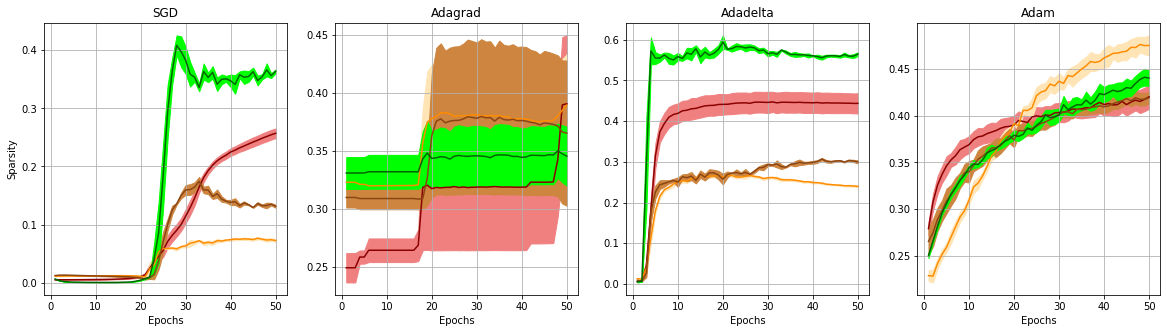}
  \includegraphics[width=1.0\textwidth]{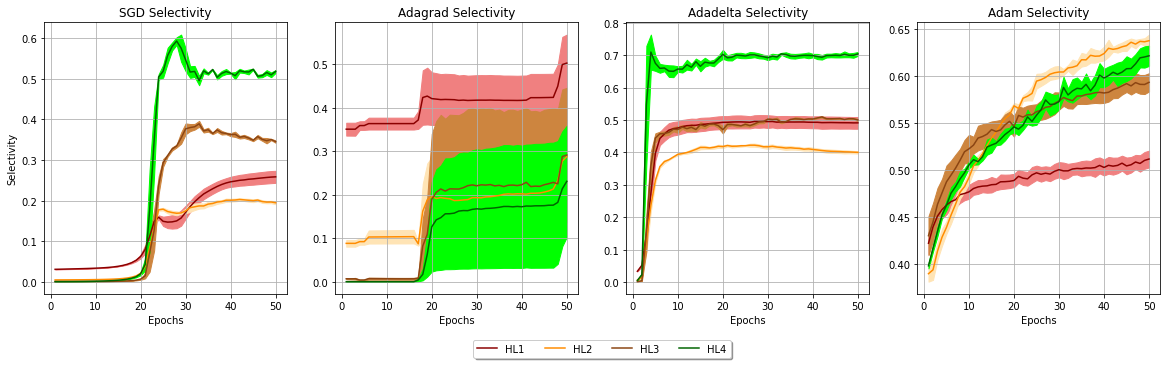}
  \caption{Test accuracy, sparsity, and selectivity of each hidden layer. There are 4 hidden layers. Data represented as mean and standard error (n=3)}
  \label{hl_eachSM}
\end{figure}

\begin{figure}[h!]
  \centering
  \includegraphics[width=1.0\textwidth]{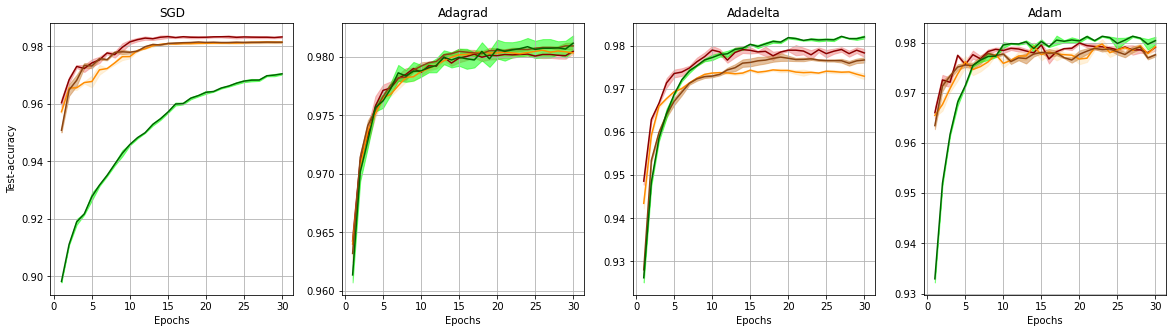}
  \includegraphics[width=1.0\textwidth]{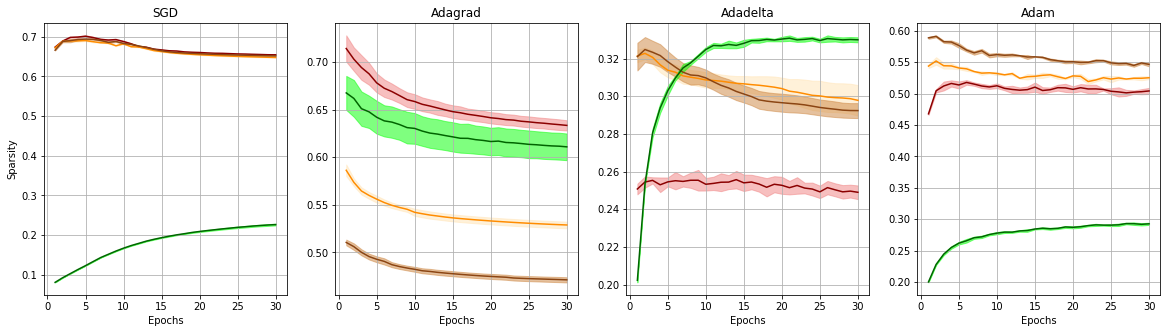}
  \includegraphics[width=1.0\textwidth]{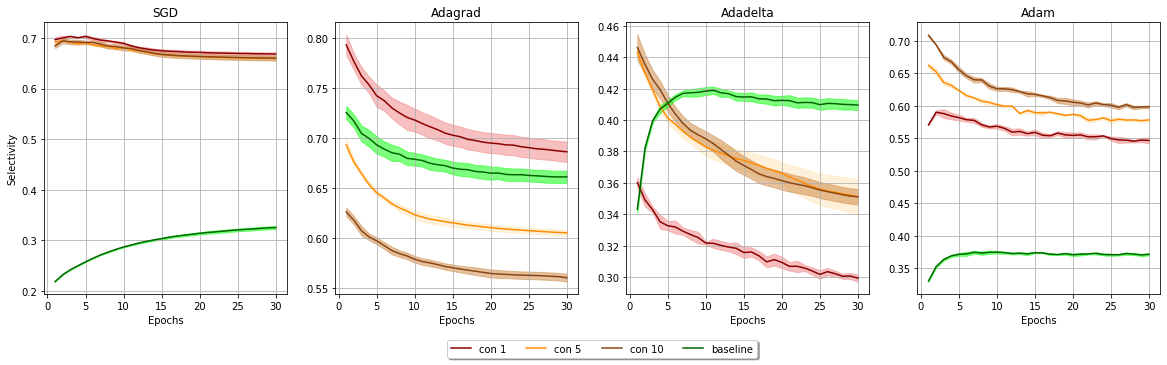}
  \caption{Test accuracy, sparsity, and selectivity when the batch size is set to 1 with the $x$ same consecutive digits ($x$ = 5, 10). See \ref{extensionofvaryingbatchsize} for the experimental details. Data represented as mean and standard error (n=3)}
  \label{batchsize1allSM}
\end{figure}

\begin{figure}[h!]
  \centering
  \includegraphics[width=1.0\textwidth]{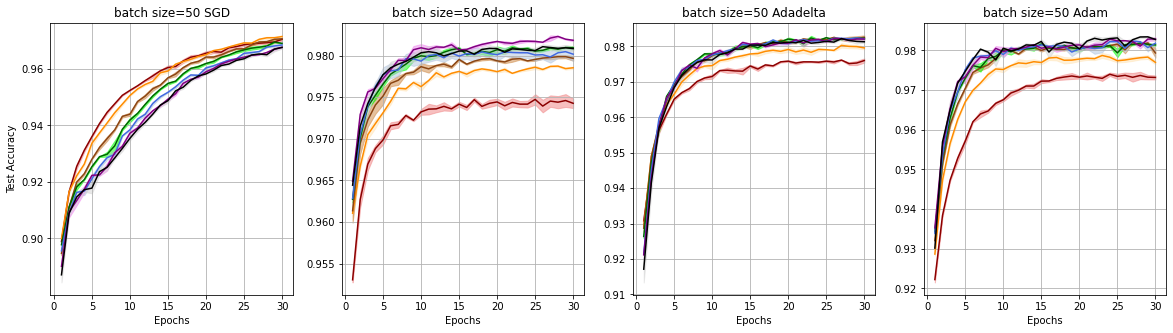}
  \includegraphics[width=1.0\textwidth]{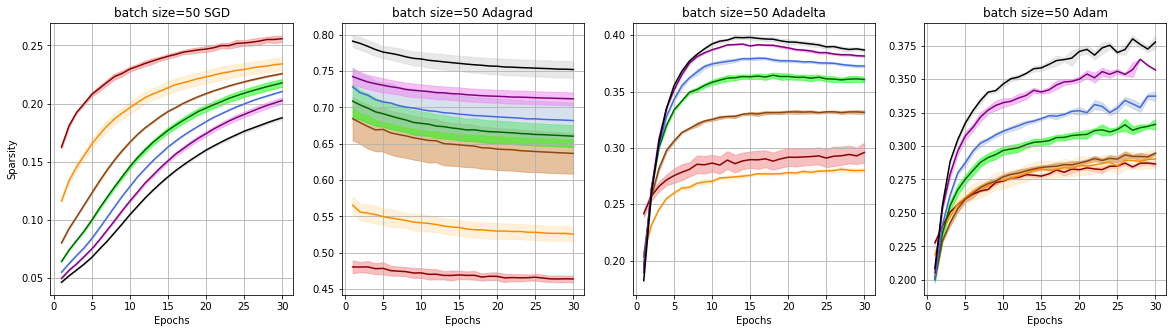}
  \includegraphics[width=1.0\textwidth]{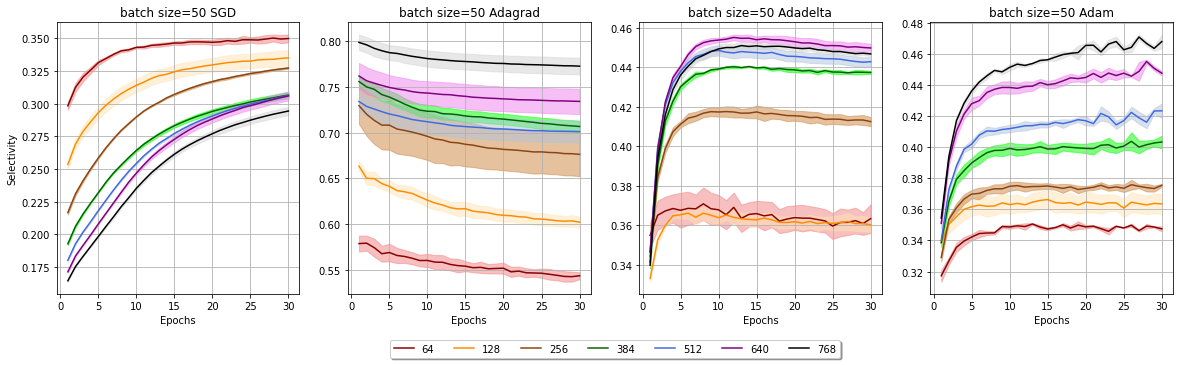}
  \caption{Test accuracy, sparsity, and selectivity when varying the number of neurons in the single hidden layer. Data represented as mean and standard error (n=3)}
  \label{number of neurons b=50}
\end{figure}

\begin{figure}[h!]
  \centering
  \includegraphics[width=1.0\textwidth]{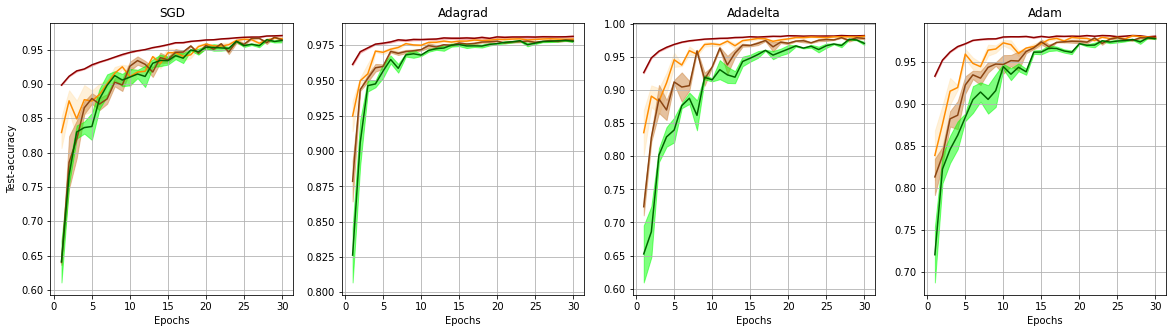}
  \includegraphics[width=1.0\textwidth]{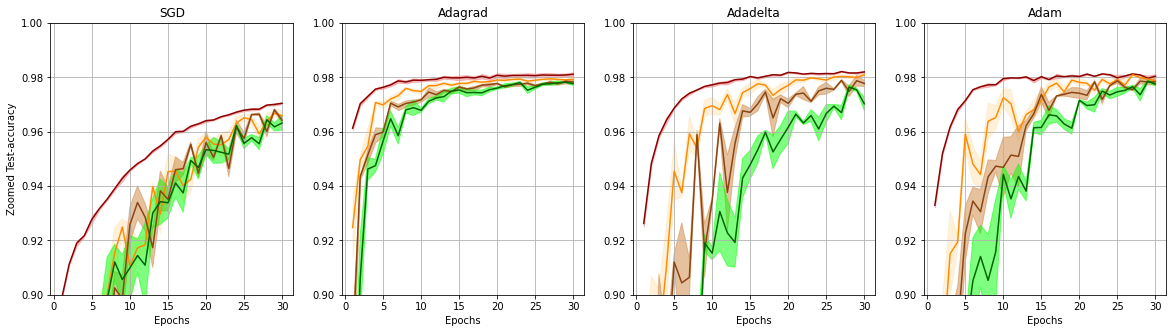}
  \includegraphics[width=1.0\textwidth]{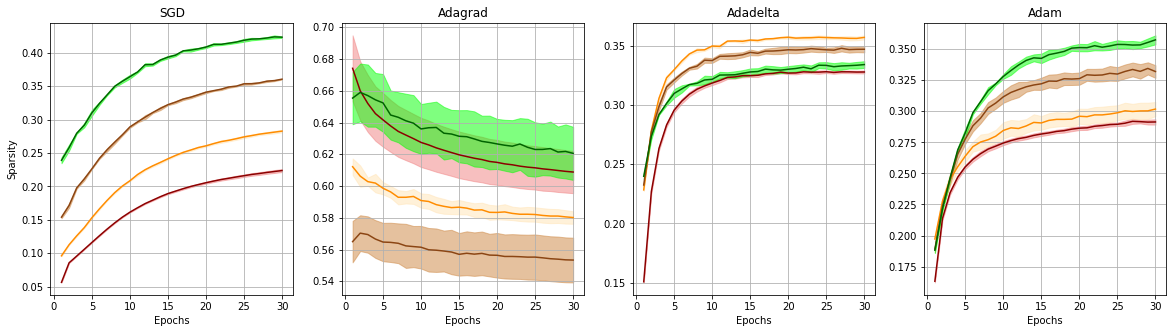}
  \includegraphics[width=1.0\textwidth]{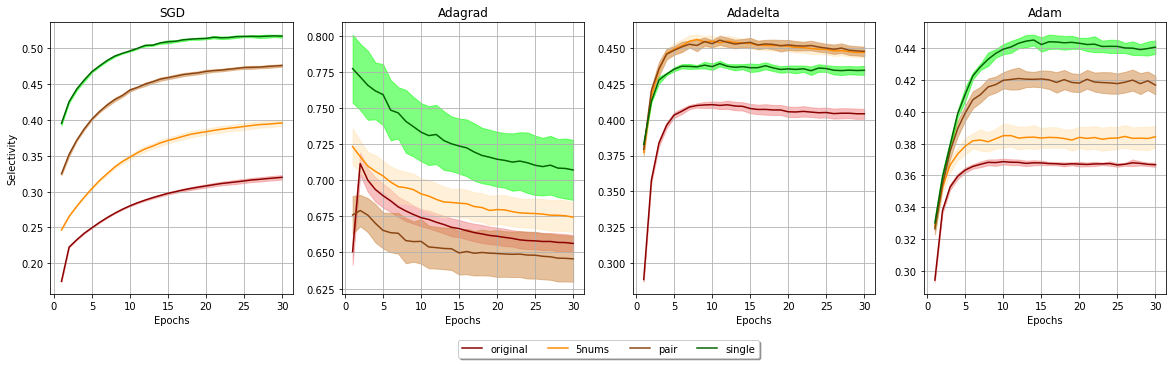}
  \caption{Test accuracy, sparsity, and selectivity when varying class diversity in a batch. See \ref{singlepair5numsexperimentaldetails} for detailed experimental details. Data represented as mean and standard error (n=3)}
  \label{singlepair5numsSM}
\end{figure}

\begin{figure}[h!]
  \centering
  \includegraphics[width=1.0\textwidth]{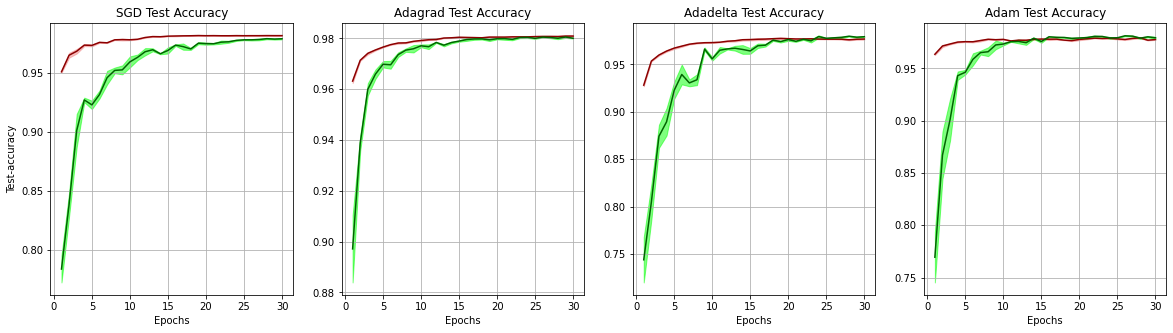}
  \includegraphics[width=1.0\textwidth]{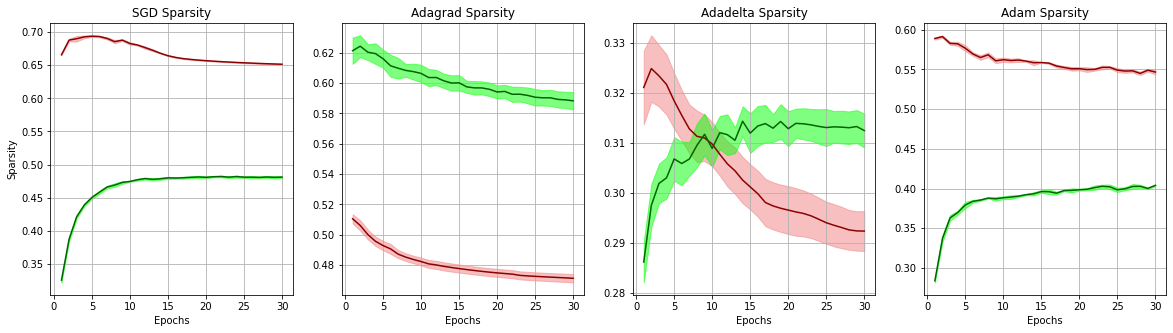}
  \includegraphics[width=1.0\textwidth]{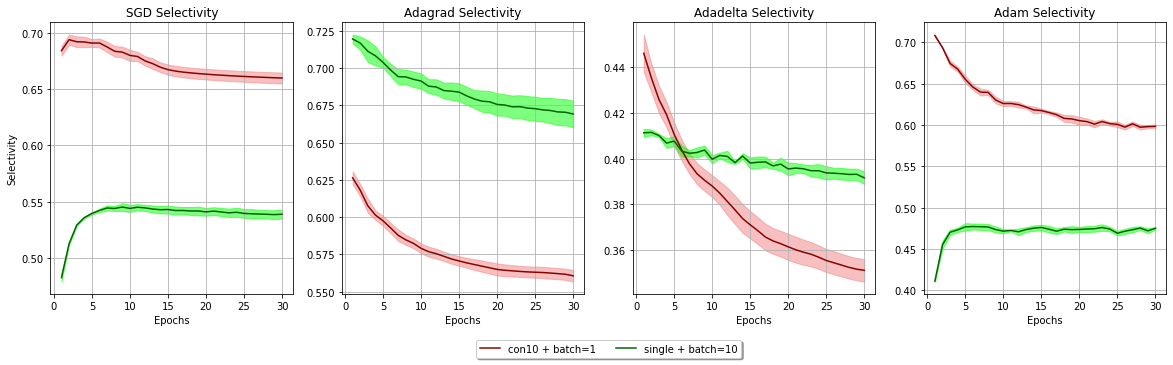}
  \caption{Test accuracy, sparsity, and selectivity comparing \textit{single+batch10} and \textit{con10+batch1}. Data represented as mean and standard error (n=3)}
  \label{single_batch10_vs_sub10_batch1_SM}
\end{figure}

%
%

\backmatter

\end{document}